\documentclass{article}

\usepackage[preprint]{neurips_2025}

\usepackage[utf8]{inputenc} 
\usepackage[T1]{fontenc}    
\usepackage{hyperref}       
\usepackage{amsfonts}       
\usepackage{nicefrac}       
\usepackage{microtype}      
\usepackage{xcolor}         
\usepackage[utf8]{inputenc}
\usepackage{longtable}
\usepackage{tabularx}
\usepackage{graphicx}
\usepackage{multirow}
\usepackage{amsfonts}
\usepackage{tikz}
\usepackage{url}            
\usepackage{booktabs}       

\usepackage[edges]{forest}
\definecolor{hidden-draw}{RGB}{20,68,106}
\definecolor{hidden-pink}{RGB}{255,245,247}
\usepackage{amsmath}
\usepackage{setspace}
\usepackage{cleveref} 
\usepackage{makecell}
\usepackage{colortbl}
\usepackage{float}
\usepackage{longtable}
\usepackage{tabularx}
\title{A Survey on Large Language Model Benchmarks}

\vspace{-3em}
\author{
  Shiwen Ni\textsuperscript{1}, Guhong Chen\textsuperscript{1, 2}, Shuaimin Li\textsuperscript{1}, Xuanang Chen\textsuperscript{9}, Siyi Li\textsuperscript{1, 4}, Bingli Wang\textsuperscript{6}\\
  \textbf{Qiyao Wang\textsuperscript{1, 3}, Xingjian Wang\textsuperscript{5}, Yifan Zhang\textsuperscript{7}, Liyang Fan\textsuperscript{8}}\\ \textbf{Chengming Li\textsuperscript{10}, Ruifeng Xu\textsuperscript{11}, Le Sun\textsuperscript{9}, Min Yang\textsuperscript{1, 12, *}} \\
  \textsuperscript{1}Shenzhen Key Laboratory for High Performance Data Mining, \\Shenzhen Institutes of Advanced Technology, Chinese Academy of Sciences \\
  \textsuperscript{2}Southern University of Science and Technology 
  \textsuperscript{3}University of Chinese Academy of Sciences \\
  \textsuperscript{4}University of Science and Technology of China
  \textsuperscript{5}Shanghai University of Electric Power \\
  \textsuperscript{6}Shanghai AI Lab 
  \textsuperscript{7}South China University of Technology
  \textsuperscript{8}Shenzhen University  \\
  \textsuperscript{9}Institute of Software, Chinese Academy of Sciences 
  \textsuperscript{10}Shenzhen MSU-BIT University \\
  \textsuperscript{11}Harbin Institute of Technology, Shenzhen  
  \textsuperscript{12}Shenzhen University of Advanced Technology
  \\
}

\begin{document}

\maketitle
\vspace{-2.5em}
\begin{figure*}[h]  %
    \centering
    \includegraphics[width=0.99\textwidth]{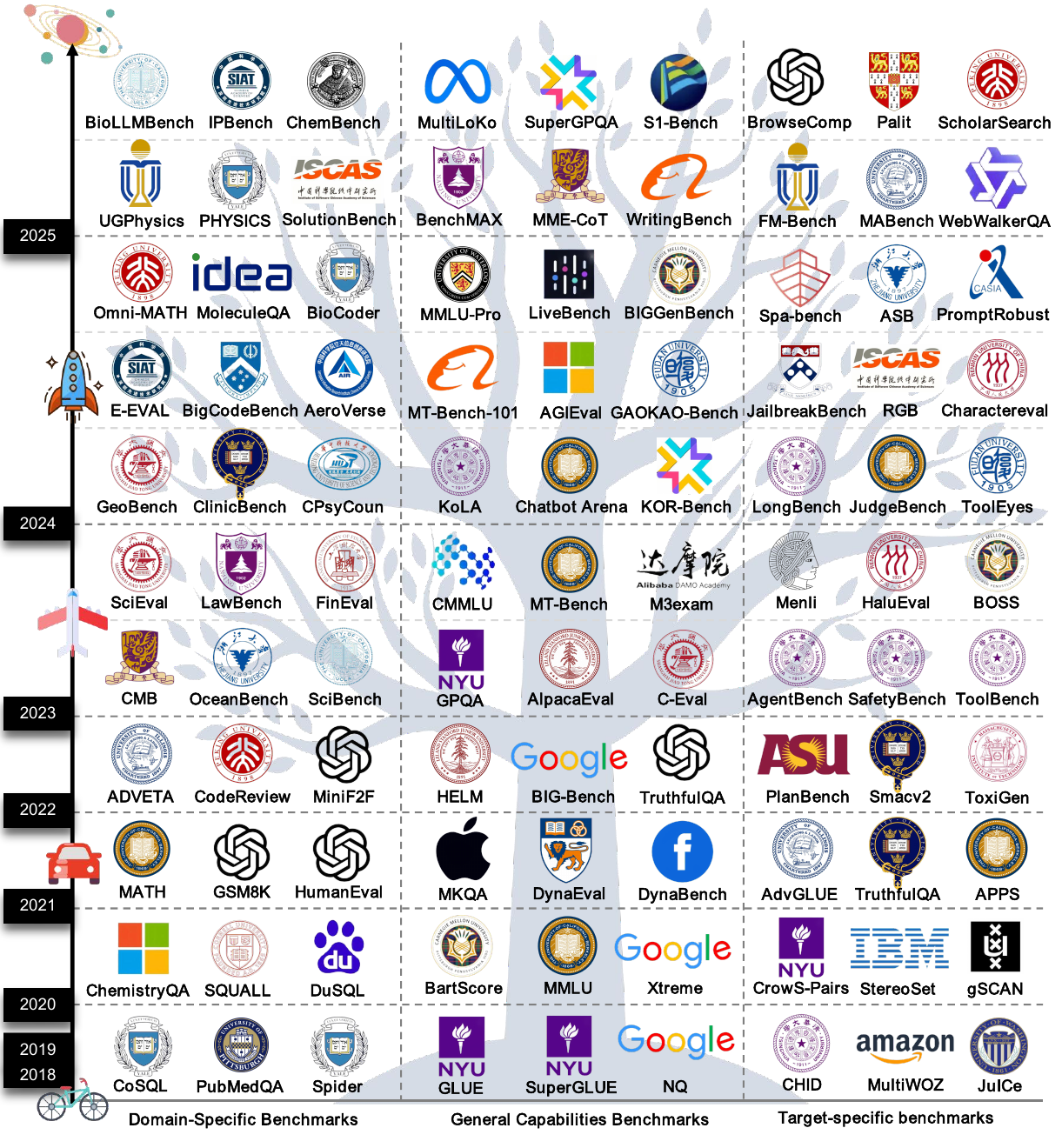}  
    \caption{A timeline of representative LLM benchmarks.}
    \label{fig:example}
    \vspace{-1.5em}
\end{figure*}

\newpage
\tableofcontents
\newpage
\begin{abstract}
In recent years, with the rapid development of the depth and breadth of large language models' capabilities, various corresponding evaluation benchmarks have been emerging in increasing numbers. As a quantitative assessment tool for model performance, benchmarks are not only a core means to measure model capabilities but also a key element in guiding the direction of model development and promoting technological innovation. We systematically review the current status and development of large language model benchmarks for the first time, categorizing 283 representative benchmarks into three categories: general capabilities, domain-specific, and target-specific. General capability benchmarks cover aspects such as core linguistics, knowledge, and reasoning; domain-specific benchmarks focus on fields like natural sciences, humanities and social sciences, and engineering technology; target-specific benchmarks pay attention to risks, reliability, agents, etc. We point out that current benchmarks have problems such as inflated scores caused by data contamination, unfair evaluation due to cultural and linguistic biases, and lack of evaluation on process credibility and dynamic environments, and provide a referable design paradigm for future benchmark innovation.
\end{abstract}
\section{Introduction} 
Since the Transformer \citep{vaswani2017attention} architecture was introduced in 2017, large language models (LLMs) have launched a revolutionary wave in the field of Artificial Intelligence (AI) with their powerful natural language processing capabilities. From basic natural language understanding and text generation tasks to complex logical reasoning and intelligent body interactions, LLMs continue to expand the boundaries of AI and reshape the human-computer interaction paradigm and information processing model.
With the successive introduction of GPT series \citep{radford2018improving,radford2019language,gpt4}, LLaMA series \citep{llama,llama2,llama3}, Qwen series \citep{qwen,qwen2,qwen3}, and other models, LLMs have widely penetrated into intelligent customer service, content creation, education, medical care, law and other fields, and have become the core driving force to promote the development of the digital economy and the intelligent transformation of society.

With the acceleration of the iteration of the LLM technology, it is urgent to establish a scientific and comprehensive evaluation system; Benchmarks, as a quantitative assessment of model performance, are not only a core tool to measure the ability of the model, but also a key element to guide the direction of model and promote technological innovation. Through benchmarks, researchers can objectively compare the strengths and weaknesses of different models, accurately locate technical bottlenecks, and provide data support for algorithm optimization and architectural design; At the same time, standardized evaluation results can help build user trust and ensure that the models comply with the social and ethical norms in terms of security and fairness. However, compared with the earlier language model evaluation benchmarks represented by GLUE \cite{GLUE} and SuperGLUE \cite{SuperGLUE}, the number of model parameters in the LLM era has increased exponentially, the capability dimension has expanded from single task to multitask and multidomain MMLU \cite{MMLU}, GIG-bench \cite{BIG-Bench}, GPQA \cite{GPQA}, SuperGPQA \cite{SuperGPQA}, and the evaluation paradigm has shifted from fixed task to multitask and multidomain. These changes put forward higher requirements on the scientific and adaptive nature of evaluation systems.

Currently, the field of LLM evaluation still faces many challenges that need to be solved. First, data leakage \citep{ni2025training,C2LEVA} is becoming increasingly prominent, and some models are exposed to evaluation data during the training phase, leading to inflated evaluation results and fails to truly reflect the model generalization ability; second, static evaluation \cite{MMLU,Mmlu-pro} is difficult to simulate dynamic real-world scenarios and it is difficult to predict model performance when faced with new tasks and new domains. The singularity of evaluation indexes (e.g., over-reliance on accuracy rate and BLEU score) fails to comprehensively portray the complex capabilities of LLMs, and key requirements such as detection of bias and security loopholes and systematic evaluation of instruction compliance have not yet been effectively met. In addition, the high cost of arithmetic and manpower required for large-scale evaluation, and the difficulty of task design to cover the complexity of the real world are serious constraints to the healthy development of LLMs. Figure \ref{fig:example} shows a timeline of representative LLM benchmarks, illustrating this rapid evolution.

This paper is the first to conduct a systematic review and prospective analysis focusing on LLM benchmarks, with its contributions summarized as follows:

\begin{enumerate}
    \item For the first time, 283 LLM benchmarks are analyzed and summarized under three categories: General Capabilities Benchmarks, Domain-Specific Benchmarks, and Target-specific Benchmarks.
    \item This paper examines the design motivations and limitations of each benchmark from multiple perspectives, including data sources, data formats, data volume, evaluation methods, and evaluation metrics, providing a directly referable design paradigm for subsequent benchmark innovation.
    \item We point out three major issues faced by current LLM benchmarks: inflated scores caused by data contamination; unfair evaluation due to cultural and linguistic biases; and the lack of evaluation on "process credibility" and "dynamic environments".
\end{enumerate}

\section{Background}
\subsection{Large Language Models}
Research on language models dates back to Shannon in the 1950s, who pioneered modeling human language with information theory using n-gram models~\cite{Shannonlanguagemodel}. Its evolution has gone through several stages: statistical language models (e.g., n-gram models relying on co-occurrence statistics and independence assumptions~\cite{jelinek1998statistical,stolcke2002srilm}), neural language models (utilizing distributed representations and architectures like recurrent neural networks~\cite{mikolov2010recurrent,KombrinkMKB11,BengioDV00}, with works such as word2vec advancing representation learning~\cite{mikolov2013distributed,mikolov2013efficient}), and subsequently pretrained language models (PLMs).

Pretrained language models learn context-aware representations from large unlabeled corpora for downstream fine-tuning. ELMo introduced bidirectional LSTM pretraining for dynamic embeddings~\cite{peters-etal-2018-deep}. The Transformer architecture, with its self-attention mechanism, became foundational for large-scale PLMs~\cite{vaswani2017attention}. Based on this, models like BERT~\cite{devlin-etal-2019-bert}, GPT/GPT-2~\cite{radford2018improving,radford2019language}, BART~\cite{lewis2019bart}, and T5~\cite{raffel2020exploring} emerged, following the "pretraining and fine-tuning" paradigm, with subsequent refinements (e.g., RoBERTa~\cite{liu2019roberta}).

Large language models (LLMs) developed from PLMs, driven by scaling laws that link increased parameters and data to improved performance~\cite{kaplan2020scaling}. With parameter counts growing to billions/trillions, LLMs exhibit emergent capabilities like few-shot learning and in-context learning~\cite{wei2022emergent}. The ecosystem includes proprietary models (e.g., OpenAI's ChatGPT, GPT-4~\cite{gpt4}; Anthropic's Claude; Google's Gemini~\cite{team2023gemini}) and open-source ones (e.g., Meta’s LLaMA series~\cite{llama}; Alibaba’s Qwen series~\cite{qwen,qwen2,qwen3}), excelling in diverse tasks from dialogue to multimodal reasoning.

\tikzstyle{my-box}=[
    rectangle,
    draw=hidden-draw,
    rounded corners,
    text opacity=1,
    minimum height=1.5em,
    minimum width=5em,
    inner sep=2pt,
    align=center,
    fill opacity=.5,
    line width=0.8pt,
]
\tikzstyle{leaf}=[my-box, minimum height=1.5em,
    fill=hidden-pink!80, text=black, align=left,font=\small,
    inner xsep=2pt,
    inner ysep=4pt,
    line width=0.8pt,
]

\begin{figure*}[t]
    \centering
    \resizebox{\textwidth}{!}{
        \begin{forest}
            forked edges,
            for tree={
                grow=east,
                reversed=true,
                anchor=base west,
                parent anchor=east,
                child anchor=west,
                base=center,
                font=\large,
                rectangle,
                draw=hidden-draw,
                rounded corners,
                align=left,
                text centered,
                minimum width=13em,
                edge+={darkgray, line width=1pt},
                s sep=6pt, 
                l sep=10pt,
                inner xsep=5pt, 
                inner ysep=5pt, 
                line width=0.8pt,
                ver/.style={rotate=90, child anchor=north, parent anchor=south, anchor=center},
            },
            where level=1{text width=15em,font=\normalsize,}{},
            where level=2{text width=14em,font=\normalsize,}{},
            where level=3{text width=50em,}{}, 
            [
                \textbf{Benchmarks of LLMs}, ver, fill=yellow!20
                [
                    \textbf{General Capabilities Benchmarks}, fill=green!15
                    [
                        \textbf{Linguistic Core}, fill=green!8
                        [
                            1. \textbf{NLU:} GLUE \cite{GLUE}; SuperGLUE \cite{SuperGLUE}; CLUE \cite{CLUE}; TriviaQA \cite{TriviaQA}; NaturalQuestions \cite{NaturalQuestions} \\
                            2. \textbf{Commonsense:} HellaSwag \cite{HellaSwag}; WinoGrande \cite{WinoGrande}; ProtoQA \cite{ProtoQA} \\
                            3. \textbf{Generation:} BERTScore \cite{BERTScore}; Bartscore \cite{Bartscore}; BLEURT \cite{BLEURT}; BiGGenBench \cite{BiGGenBench}; AlpacaEval \cite{AlpacaEval}; ROUGE \cite{ROUGE}; BLEU \cite{BLEU} \\
                            4. \textbf{Dialogue:} MT-Bench \cite{MT-Bench}; MT-Bench-101 \cite{MT-Bench-101}; DynaEval \cite{DynaEval}; MDIA \cite{MDIA}; FaithDial \cite{FaithDial} \\
                            5. \textbf{Multilingual:} Xtreme \cite{Xtreme} \\
                            6. \textbf{Holistic:} HELM \cite{HELM}; BIG-Bench \cite{BIG-Bench}
                            ,leaf
                        ]
                    ]
                    [
                        \textbf{Knowledge}, fill=green!8
                        [
                            1. \textbf{Comprehensive:} MMLU \cite{MMLU}; MMLU-Pro \cite{Mmlu-pro}; BIG-Bench \cite{BIG-Bench}; TriviaQA \cite{TriviaQA}; NaturalQuestions \cite{NaturalQuestions} \\
                            2. \textbf{Expert-Level:} GPQA \cite{GPQA}; SuperGPQA \cite{SuperGPQA} \\
                            3. \textbf{Exam-Based:} AGIEval \cite{Agieval}; GAOKAO-Bench \cite{GAOKAO-Bench}; M3Exam \cite{M3exam}; GAOKAO-MM \cite{GAOKAO-MM} \\
                            4. \textbf{Language-Specific:} C-Eval \cite{C-Eval}; CMMMU \cite{CMMMU} \\
                            5. \textbf{Granular \& Holistic:} KoLA \cite{KoLA}; HELM \cite{HELM}
                            ,leaf
                        ]
                    ]
                    [
                        \textbf{Reasoning}, fill=green!8
                        [
                            1. \textbf{Logical:} RuleTaker \cite{RuleTaker}; ProofWriter \cite{ProofWriter}; FOLIO \cite{FOLIO}; ReClor \cite{ReClor}; LogiQA \cite{LogiQA}; LogicBench \cite{LogicBench}; \\ \qquad SATBench \cite{SATBench}; LogicPro \cite{LogicPro}; LogicNLI \cite{LogicNLI}; Unigram-FOL \cite{Unigram-FOL}; P-FOLIO \cite{P-FOLIO}; Multi-LogiEval \cite{Multi-LogiEval}; \\ \qquad ZebraLogic \cite{ZebraLogic}; FineLogic \cite{FineLogic}; LogicAsker \cite{LogicAsker}; LogicInference \cite{LogicInference}; SimpleLogic \cite{SimpleLogic}; \\ \qquad AutoLogi \cite{AutoLogi}; PARAT \cite{PARAT}; DebateBench \cite{DebateBench}; PrOntoQA \cite{PrOntoQA}; PrOntoQA-OOD \cite{PrOntoQA-OOD}; LLM\_Compose \cite{LLM_Compose} \\
                            2. \textbf{Commonsense:} StrategyQA \cite{StrategyQA}; aNLI \cite{aNLI}; CommonGen \cite{CommonGen}; CLUTRR \cite{CLUTRR}; StructTest \cite{StructTest} \\
                            3. \textbf{Causal:} Corr2Cause \cite{Corr2Cause}; CLadder \cite{CLadder}; CRAB \cite{CRAB} \\
                            4. \textbf{Mathematical:} MathQA \cite{MathQA}; GSM-Symbolic \cite{GSM-Symbolic}; MathEval \cite{MathEval}; Mathador-LM \cite{Mathador-LM} \\
                            5. \textbf{Applied:} ARC \cite{ARC}; HotpotQA \cite{HotpotQA}; LiveBench \cite{LiveBench}; SysBench \cite{SysBench}; AR-Bench \cite{AR-Bench}; BIG-Bench Hard \cite{BIG-Bench-Hard}; \\ \qquad IOLBENCH \cite{iolbench}; ANALOGICAL \cite{ANALOGICAL}; TextGames \cite{TextGames}
                            ,leaf
                        ]
                    ]
                ]
                [
                    \textbf{Domain-Specific Benchmarks}, fill=purple!15
                    [
                        \textbf{Natural Sciences}, fill=purple!8
                        [
                            1. \textbf{Math:} GSM8K \cite{GSM8K}; MATH \cite{MATH}; Omni-MATH \cite{Omni-MATH}; FrontierMath \cite{frontiermath}; JEEBench \cite{JEEBench}; U-MATH \cite{u-math}; \\ \qquad MiniF2F \cite{minif2f}; MATH-P \cite{math-p}; MathBench \cite{mathbench}; Hardmath \cite{hardmath}; Lila \cite{lila}; HARP \cite{harp}; ASyMOB \cite{asymob} \\
                            2. \textbf{Physics:} SciBench \cite{scibench}; UGPhysics \cite{ugphysics}; SeePhys \cite{seephys}; PhysicsArena \cite{physicsarena}; PhysReason \cite{physreason}; PhysUniBench \cite{physunibench}; \\ \qquad  FEABench \cite{FEABench}; PHYSICS \cite{physics}; TPBench \cite{tpbench} \\
                            3. \textbf{Chemistry:} ChemEval \cite{chemeval}; MoleculeQA \cite{moleculeqa}; ChemSafetyBench \cite{chemsafetybench}; ChemistryQA \cite{chemistryqa}; ScholarChemQA \cite{scholarchemqa}; \\ \qquad ChemLLMBench \cite{chemllmbench}; Alchemy \cite{alchemy} \\
                            4. \textbf{Biology:} PubMedQA \cite{pubmedqa}; BioMaze \cite{biomaze}; SciAssess \cite{sciassess}; BixBench \cite{BixBench}; LAB-Bench \cite{labbench};  AutoBio \cite{autobio}; \\ \qquad BioProBench \cite{bioprobench} \\
                            5. \textbf{Cross-Disciplinary:} GPQA \cite{GPQA}; TheoremQA \cite{theoremqa}; SCITOOLBENCH \cite{sciagent}; SciReplicate-Bench \cite{SciReplicate-Bench}; \\ \qquad SciEval \cite{scieval}; LLM-SRBench \cite{LLM-SRBench}; OpenBookQA \cite{OpenBookQA}; WMDP \cite{wmdp}; Curie \cite{curie}; SciKnowEval \cite{sciknoweval}; GEO-Bench \cite{geo-bench}
                            ,leaf
                        ]
                    ]
                    [
                        \textbf{Humanities \& Social Sciences}, fill=purple!8
                        [
                            1. \textbf{Law:} LegalBench \cite{Legalbench}; LawBench \cite{LawBench}; LBOXOPEN \cite{LBOXOPEN}; LexEval \cite{LexEval}; LAiW \cite{LAiW}; CiteLaw \cite{CiteLaw}; CaseGen \cite{CaseGen} \\
                            2. \textbf{IP:} PatentEval \cite{PatentEval}; MoZIP \cite{MoZIP}; IPBench \cite{IPBench}; IPEval \cite{IPEval}; D2P \cite{D2P} \\
                            3. \textbf{Education:} E-Eval \cite{E-EVAL}; EduBench \cite{EduBench} \\
                            4. \textbf{Psychology:} CPsyExam \cite{CPsyExam}; CPsyCoun \cite{CPsyCoun}; Psycollm \cite{Psycollm}; PsychoBench \cite{PsychBench}; Psychometrics Benchmark \cite{li2024quantifying} \\
                            5. \textbf{Finance:} FinEval \cite{Fineval}; FLARE \cite{FLARE}; BBT-CFLEB \cite{BBT-CFLEB}
                            ,leaf
                        ]
                    ]
                    [
                        \textbf{Engineering \& Technology}, fill=purple!8
                        [
                            1. \textbf{Code Gen:} HumanEval \cite{HumanEval}; MBPP \cite{MBPP}; APPS \cite{APPS}; LiveCodeBench \cite{LiveCodeBench}; BigCodeBench \cite{BigCodeBench}; DS-1000 \cite{DS-1000}; \\ \qquad BioCoder \cite{BioCoder}; USACO \cite{USACO}; LiveCodeBenchPro \cite{LiveCodeBenchPro}; ClassEval \cite{ClassEval}; MMCode \cite{MMCode} \\
                            2. \textbf{Code Maint. \& Repair:} SWE-bench \cite{SWE-bench}; RepairBench \cite{RepairBench}; Debugbench \cite{Debugbench}; Condefects \cite{Condefects};  CanItEdit \cite{CanItEdit}; \\ \qquad CodeEditorBench \cite{CodeEditorBench}; COFFE \cite{COFFE}; EffiBench \cite{EffiBench} \\
                            3. \textbf{Code Understanding:} CodeXGLUE \cite{CodeXGLUE}; xCodeEval \cite{xCodeEval}; CodeQA \cite{CodeQA}; Cosqa \cite{Cosqa}; Repobench \cite{Repobench}; \\ \qquad CodeReview \cite{CodeReview} \\
                            4. \textbf{Database \& DevOps:} Spider \cite{Spider}; Dr. Spider \cite{DrSpider}; BIRD \cite{BIRD}; CoSQL \cite{CoSQL}; IaC-Eval \cite{IaC-Eval}; OpsEval \cite{OpsEval}; \\ \qquad Spider 2.0 \cite{Spider2}; SParC \cite{SParC}; DuSQL \cite{DuSQL}; NL2Bash \cite{NL2Bash}; OWL \cite{OWL-Bench}; FrontendBench \cite{FrontendBench} \\
                            5. \textbf{Hardware \& Eng.:} VerilogEval \cite{VerilogEval}; RTLLM \cite{RTLLM}; CIRCUIT \cite{CIRCUIT}; FIXME \cite{FIXME}; CADBench \cite{CADBench}; LLM4Mat-bench \cite{LLM4Mat-bench}; \\ \qquad  Aviation-Benchmark \cite{Aviation-Benchmark}; ResBench \cite{ResBench}; PICBench \cite{PICBench}; ElecBench \cite{ElecBench} \\
                            6. \textbf{Other Eng.:} MSQA \cite{MSQA}; AeroMfg-QA \cite{AeroManufacturing-QA}; RepoSpace \cite{RepoSpace}
                            ,leaf
                        ]
                    ]
                ]
                [
                    \textbf{Target-specific Benchmarks}, fill=blue!15
                    [
                        \textbf{Risk \& Reliability}, fill=blue!8
                        [
                            1. \textbf{Safety:} StereoSet \cite{StereoSet}; ToxiGen \cite{ToxiGen}; JailbreakBench \cite{JailbreakBench}; Do-Not-Answer \cite{Do-Not-Answer}; HateCheck \cite{HateCheck};  CrowS-Pairs \cite{CrowS-Pairs}; \\ \qquad HarmBench \cite{HarmBench}; ToxicChat \cite{ToxicChat}; SG-Bench \cite{SG-Bench}; AnswerCarefully \cite{AnswerCarefully};  SorryBench \cite{SorryBench}; MaliciousInstruct \cite{MaliciousInstruct}; \\ \qquad 
                            HEx-PHI \cite{HEx-PHI}; SimpleSafetyTests \cite{Simplesafetytests}; In-The-Wild Jailbreak Prompts \cite{In-The-Wild} \\
                            2. \textbf{Hallucination:} TruthfulQA \cite{TruthfulQA}; FActScore \cite{FActScore}; HaluEval \cite{HaluEval}; HaluEval2.0 \cite{HaluEval2.0}; FreshQA \cite{FreshQA}; FaithDial \cite{FaithDial}; \\ \qquad RealtimeQA \cite{RealtimeQA}; FaithBench \cite{FaithBench}; DiaHalu \cite{DiaHalu}; FactCheck-Bench \cite{FactCheckBench};  FELM \cite{FELM}; FACTOR \cite{FACTOR}; MedHallu \cite{MedHallu} \\
                            3. \textbf{Robustness:} AdvGLUE \cite{AdvGLUE}; IFEval \cite{IFEval}; PromptRobust \cite{PromptRobust}; BOSS \cite{BOSS}; CIF-Bench \cite{CIF-Bench}; RoTBench \cite{RoTBench} \\
                            4. \textbf{Data Leak:} WikiMIA \cite{WikiMIA}; C$^2$LEVA \cite{C2LEVA}; KoLA \cite{KoLA}
                            ,leaf
                        ]
                    ]
                    [
                        \textbf{Agent}, fill=blue!8
                        [
                            1. \textbf{Planning \& Control:} FlowBench \cite{FlowBench}; Mobile-Bench \cite{Mobile-Bench}; WebWalkerQA \cite{WebWalkerQA}; Robotouille \cite{Robotouille}; BrowseComp \cite{BrowseComp}; \\ \qquad LLF-Bench \cite{LLF-Bench}; Spa-Bench \cite{Spa-Bench} \\
                            2. \textbf{Multi-Agent:} MultiAgentBench \cite{MultiAgentBench}; MAgIC \cite{MAgIC}; ZSC-Eval \cite{ZSC-Eval} \\
                            3. \textbf{Integrated \& Holistic:} GAIA \cite{GAIA}; AgentBench \cite{AgentBench}; AgentBoard \cite{AgentBoard}; AgentQuest \cite{AgentQuest};  TravelPlanner \cite{TravelPlanner}; \\ \qquad SmartPlay \cite{SmartPlay}; CharacterEval \cite{CharacterEval}; BALROG \cite{BALROG}; Embodied Agent Interface \cite{EmbodiedAgentInterface}; $\tau$-bench \cite{tau-bench}; ColBench \cite{ColBench} \\
                            4. \textbf{Domain-Specific:} OSWorld \cite{OSWorld}; ScienceAgentBench \cite{ScienceAgentBench}; AgentClinic \cite{AgentClinic}; MLGym-Bench \cite{MLGym-Bench};  InvestorBench \cite{InvestorBench}; \\ \qquad SciReplicate-Bench \cite{SciReplicate-Bench}; BixBench \cite{BixBench}; CourtBench \cite{CourtBench}; Tapilot-Crossing \cite{Tapilot-Crossing}; TheAgentCompany \cite{TheAgentCompany} \\
                            5. \textbf{Safety:} AgentHarm \cite{AgentHarm}; SafeAgentBench \cite{SafeAgentBench}; R-Judge \cite{R-Judge}; ASB \cite{ASB}
                            ,leaf
                        ]
                    ]
                    [
                        \textbf{Others}, fill=blue!8
                        [
                            PET-Bench \cite{pet-bech}; TP-RAG \cite{tp-rag}; FLUB \cite{FLUB}; CDEval \cite{cdeval}; NORMAD-ETI \cite{normad}; JudgeBench \cite{JudgeBench}; EmotionQueen\cite{emotionqueen}; \\ OR-Bench\cite{orbench}; SocialStigmaQA\cite{SocialStigmaQA}; ShoppingMMLU\cite{ShoppingMMLU}; LLM-Evolve \cite{LLM-Evolve}; DOCBENCH \cite{DOCBENCH}; VisEval \cite{VisEval}; SUC~\cite{tablellm}; \\ ROUTERBENCH~\cite{RouterBench}; GAME COMPETITIONS~\cite{Grid-Based-Game}; RTLLM 2.0~\cite{OpenLLM-RTL}; AD-LLM~\cite{AD-LLM}; ZIQI-Eval~\cite{ZIQI-Eval}
                            ,leaf
                        ]
                    ]
                ]
            ]
        \end{forest}
    }
    \vspace{-2mm}
    \caption{A taxonomy of representative benchmarks for Large Language Models, categorized by their primary evaluation focus.}
    \label{fig:taxonomy_of_benchmarks}
    \vspace{-2mm}
\end{figure*}
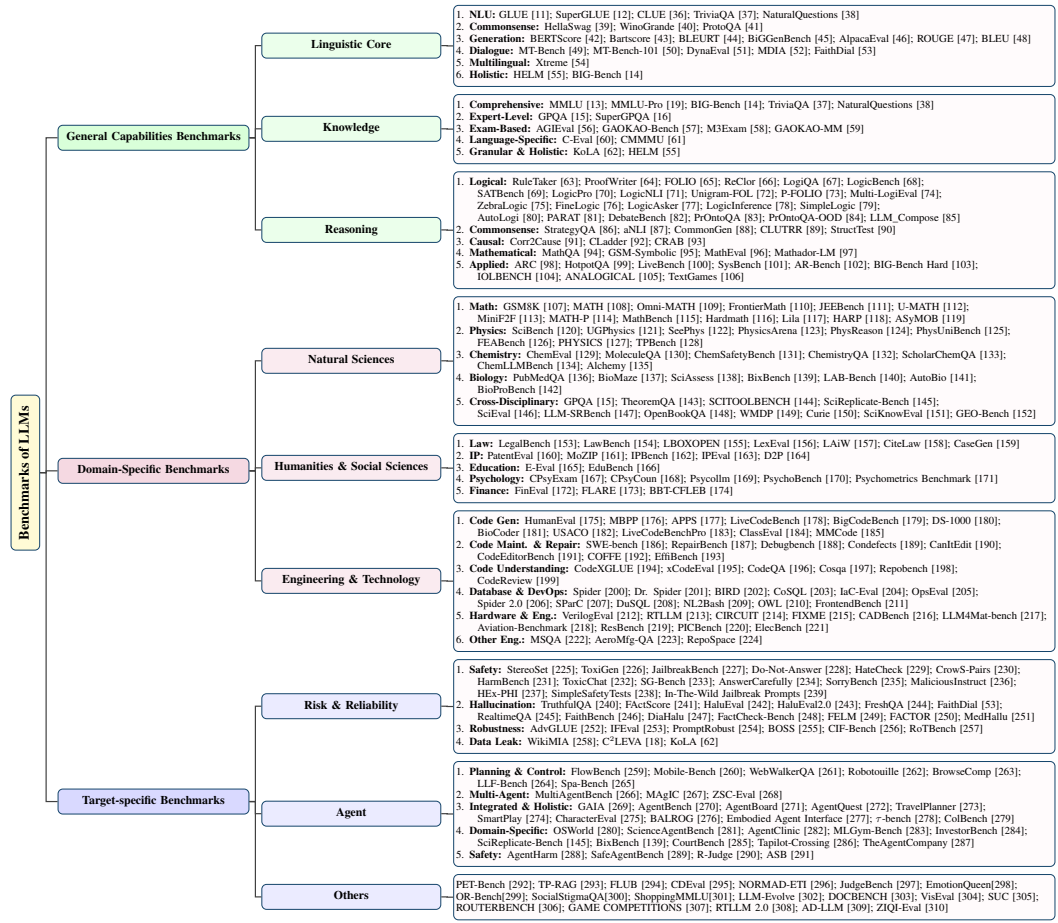

\subsection{LLM Benchmarks}
The rapid advancement of large language models (LLMs) has fundamentally reshaped the landscape of natural language processing. As LLMs grow in scale—from millions to billions and now trillions of parameters—their emergent capabilities, including complex reasoning, instruction following, multi-turn dialogue, and tool usage, have presented unprecedented opportunities and challenges. In parallel with these developments, the design and evolution of benchmarks have become essential to accurately measure, compare, and guide the progress of LLMs. Benchmarks and LLMs are not independent trajectories; instead, their evolution is deeply intertwined, forming a mutually reinforcing cycle that continuously pushes the boundaries of the field.

In the early stages of language model development, benchmarks such as GLUE~\cite{GLUE}, BERTScore~\cite{BERTScore} and SuperGLUE~\cite{SuperGLUE} played a crucial role in driving research progress. These benchmarks primarily focused on natural language understanding (NLU) through relatively small-scale, single-task evaluations. However, as LLMs rapidly scaled up in size and began to exhibit emergent generalization abilities. In response, a new wave of LLM-specific benchmarks has emerged, such as MMLU~\cite{MMLU}, BIG-bench~\cite{BIG-Bench}, HELM~\cite{HELM}, AGIEval~\cite{Agieval}, GPQA~\cite{GPQA}. These benchmarks aim to assess a wider range of capabilities, including reasoning, factual knowledge, robustness, multilingual understanding, and generalization to unseen tasks. Moreover, many of them are designed to evaluate LLMs in zero-shot or few-shot settings, aligning more closely with how these models are used in practice.

In this survey, we present a comprehensive review of LLM benchmarks, delving into their design principles, coverage scope, inherent limitations, and emerging trends. Our objective is to crystallize the current landscape of LLM evaluation and offer actionable insights to inform the development of benchmarking strategies tailored for next-generation language models. Figure \ref{fig:taxonomy_of_benchmarks} provides a detailed taxonomy of these benchmarks, which serves as the organizational structure for the remainder of this paper.

\section{General Capabilities Benchmarks}
\subsection{Linguistic Core} 
The evolution of linguistic capability benchmarks embodies a continuous arms race between model advancement and evaluation methodology. This progression is driven by the core challenge of measuring generalized linguistic competence, moving beyond surface-level pattern matching to assess deeper aspects of syntax, semantics, and pragmatics. This section chronicles how benchmarks evolved from fragmented task evaluations to dynamic, multilingual ecosystems, revealing fundamental shifts in how we define and quantify linguistic intelligence. A summary of representative benchmarks is provided in Table \ref{tab:linguistic_core_benchmarks}.

\begin{table}[h!]
\centering
\renewcommand{\arraystretch}{1}
\resizebox{\textwidth}{!}{
\begin{tabular}{llllllllll}
\hline
Benchmark & Focus & Language & Source & Data Type & Eval. & Indicators & Amount & Method & Citations \\ \hline
GLUE \cite{GLUE} & \begin{tabular}[t]{@{}l@{}}Natural Language\\ Understanding\end{tabular} & Monolingual & Hybrid & Hybrid & AE & Accuracy & 415,354 & No & 8607 \\
SuperGLUE \cite{SuperGLUE} & \begin{tabular}[t]{@{}l@{}}Natural Language\\ Understanding\end{tabular} & Monolingual & Hybrid & Hybrid & AE & Accuracy & 20,483 & No & 2662 \\
HellaSwag \cite{HellaSwag} & \begin{tabular}[t]{@{}l@{}}Commonsense\\ Inference\end{tabular} & Monolingual & Web & MCQA & AE & Accuracy & 10,004 & Yes & 2475 \\
WinoGrande \cite{WinoGrande} & \begin{tabular}[t]{@{}l@{}}Commonsense\\ Inference\end{tabular} & Monolingual & Hybrid & MCQA & AE & Accuracy & 44,000 & Yes & 2025 \\
BERTScore \cite{BERTScore} & \begin{tabular}[t]{@{}l@{}}Text\\ Generation\end{tabular} & Multilingual & Open Datasets & Generation & LLM & \begin{tabular}[t]{@{}l@{}}Precision,\\ Recall, F1\end{tabular} & 3 datasets & Yes & 6977 \\
CLUE \cite{CLUE} & \begin{tabular}[t]{@{}l@{}}Natural Language\\ Understanding\end{tabular} & Monolingual & Hybrid & Hybrid & ME & \begin{tabular}[t]{@{}l@{}}Accuracy,\\ EM\end{tabular} & 900k+ & No & 397 \\
Xtreme \cite{Xtreme} & \begin{tabular}[t]{@{}l@{}}Multilingual\\ Performance\end{tabular} & Multilingual & Hybrid & Hybrid & AE & \begin{tabular}[t]{@{}l@{}}Accuracy,\\ EM, F1\end{tabular} & 3 datasets & No & 1017 \\
Bartscore \cite{Bartscore} & \begin{tabular}[t]{@{}l@{}}Text\\ Generation\end{tabular} & Multilingual & Open Datasets & Generation & LLM & BartScore & 16 datasets & Yes & 906 \\
DynaEval \cite{DynaEval} & Dialogue & Monolingual & Open Datasets & Classification & AE & \begin{tabular}[t]{@{}l@{}}Accuracy,\\ F1\end{tabular} & 50k+ & Yes & 80 \\
HELM \cite{HELM} & \begin{tabular}[t]{@{}l@{}}Holistic\\ Capability\end{tabular} & Multilingual & Hybrid & Hybrid & AE & \begin{tabular}[t]{@{}l@{}}Accuracy,\\ 6 Designed Metrics\end{tabular} & 17,431,479 & No & 1507 \\
MT-Bench \cite{MT-Bench} & \begin{tabular}[t]{@{}l@{}}Multi-Turn\\ Dialogues\end{tabular} & Monolingual & Manual Design & Generation & LLM & 8 Designed Metrics & 80 & Yes & 4033 \\
MDIA \cite{MDIA} & \begin{tabular}[t]{@{}l@{}}Multilingual\\ Performance\end{tabular} & Multilingual & Web & Generation & ME & \begin{tabular}[t]{@{}l@{}}4 Automated Metrics\\ \& SSA\end{tabular} & 380,914 & No & 17 \\
BIG-Bench \cite{BIG-Bench}& \begin{tabular}[t]{@{}l@{}}Holistic\\ Capability\end{tabular} & Multilingual & Manual Design & Hybrid & ME & \begin{tabular}[t]{@{}l@{}}EM, MC\_Acc,\\ Breakthroughness, etc\end{tabular} & 200k+ & No & 1628 \\
BiGGenBench \cite{BiGGenBench} & \begin{tabular}[t]{@{}l@{}}Text\\ Generation\end{tabular} & Multilingual & Hybrid & Generation & ME & \begin{tabular}[t]{@{}l@{}}Instr. Follow,\\ Grounding, etc\end{tabular} & 765 & No & 7 \\ \hline
\end{tabular}
}
\caption{Summary of representative linguistic-core benchmarks. Evaluation methods are abbreviated as MCQA (Multiple Choice Question Answering), AE (Automated Evaluation), ME (Mixed Evaluation), SSA (Sensibleness and Specificity Average), MC\_ACC (Multiple-Choice Accuracy). The 'Method' column indicates if the paper proposed a new methodology (Yes/No).}
\label{tab:linguistic_core_benchmarks}
\end{table}

\subsubsection{The Evolution of Linguistic Benchmarks} \label{subsubsec:evolution}

\textbf{Phase 1: The Fragmentation Crisis and GLUE's Unification (2018)} 

Early natural language understanding (NLU) systems excelled at narrow tasks but failed to transfer knowledge across domains—a critical flaw for real-world applicability. GLUE \cite{GLUE}, introduced in 2018, was a pivotal development, confronting this by integrating 9 diverse English NLU tasks (e.g., sentiment analysis, textual entailment) under a unified framework. Its diagnostic suite was crucial, exposing models' reliance on spurious statistical cues and lexical overlap rather than an understanding of syntactic structure or semantic roles. By including limited-data subtasks, GLUE incentivized the development of models that could build more robust and transferable linguistic representations, establishing multi-task evaluation as the new paradigm.

\textbf{Phase 2: The Adversarial Turn and Deeper Linguistic Probes (2019)}  

BERT's rapid domination of GLUE \cite{GLUE} (surpassing human performance) revealed a deeper crisis: benchmarks were vulnerable to dataset-specific biases. SuperGLUE \cite{SuperGLUE} responded with harder tasks requiring complex reasoning, but the field soon uncovered models' tendency to exploit annotation artifacts. This spurred a wave of adversarially constructed benchmarks.

Benchmarks like HellaSwag \cite{HellaSwag} were designed to be difficult for models yet trivial for humans by generating distractors that were semantically plausible but pragmatically absurd, directly probing commonsense and script knowledge. Concurrently, WinoGrande \cite{WinoGrande} used the AFLITE algorithm to de-bias 44,000 pronoun disambiguation problems. This forced models to properly handle anaphora and perform true coreference resolution—a fundamental syntactic-semantic challenge—rather than relying on word-association shortcuts. These innovations redefined benchmarks as active adversaries, dynamically evolving to test for deeper linguistic phenomena beyond surface patterns.

\textbf{Phase 3: Beyond Linguistic Hegemony (2020)}  

The anglophone focus of GLUE \cite{GLUE} and SuperGLUE \cite{SuperGLUE} limited the assessment of models' ability to generalize across languages with different structural properties. CLUE \cite{CLUE} was a significant first step for Chinese NLU. Xtreme \cite{Xtreme} dramatically expanded this effort to 40 languages across 12 language families, systematically testing generalization across diverse typological properties (e.g., morphology, word order). The benchmark revealed significant performance degradation when models were transferred from high-resource, analytically-structured languages (like English) to morphologically rich or agglutinative ones. This "multilingual awakening" culminated in benchmarks like MDIA \cite{MDIA}, which extended dialogue evaluation to 46 languages, emphasizing the need to evaluate models on a wide range of morphosyntactic and cultural contexts.

\textbf{Phase 4: The Generation Paradigm Shift (2019–2021)}  

With the rise of generative models, metrics based on n-gram overlap like BLEU \cite{BLEU} and ROUGE \cite{ROUGE} proved inadequate, as they fail to capture semantic equivalence. The field responded with a new class of semantic-aware metrics. BERTScore \cite{BERTScore} leveraged contextual embeddings to measure semantic similarity, while BLEURT \cite{BLEURT} trained a regression model on 6.5M synthetically perturbed sentence pairs to better align with human judgments of quality. Bartscore \cite{Bartscore} reframed evaluation as a conditional language modeling task, directly assessing the probability of a reference given a generated output, thus aligning the metric with the model's pre-training objective. Concurrently, dialogue evaluation advanced from turn-level metrics to other more creative evaluation methods. DynaEval \cite{DynaEval}, for instance, used graph-based modeling to capture the coherence and logical flow of a conversation, assessing dependencies between utterances. 

\textbf{Phase 5: The Holistic Era and Fine-Grained Assessment (2022–Present)}  

Static benchmarks struggled to keep pace with the rapidly expanding linguistic capabilities of LLMs. In response, HELM \cite{HELM} introduced a "living benchmark" concept, dynamically integrating emergent linguistic dimensions—from cross-lingual robustness to toxicity detection—through continuous scenario expansion. This fluidity found complement in BIG-Bench \cite{BIG-Bench}, where crowdsourced frontier tasks (204 challenges co-created by 442 researchers) deliberately targeted capabilities beyond contemporary models' reach, probing complex abilities like multi-step reasoning, metaphor interpretation, and theory of mind.

Concurrently, the LLM-as-Judge revolution redefined open-ended evaluation. MT-Bench \cite{MT-Bench} and MT-Bench-101 \cite{MT-Bench-101} leveraged GPT-4 to score open-ended dialogues along dimensions like perceptivity and adaptability, achieving high correlation with human judgments. BiGGenBench \cite{BiGGenBench} pushed further, assigning instance-specific criteria (e.g., "Evaluate safety in this medical advice context") to overcome the limitations of coarse-grained metrics by enabling context-sensitive evaluation, a cornerstone of pragmatics. 

This development forged an adaptive evaluation ecosystem that treats linguistic capability not as fixed competencies but as evolving phenotypes.

\subsubsection{Cross-Cutting Design Innovations} \label{subsubsec:innovations}

Benchmark evolution reveals several tectonic shifts. 

\textbf{From Static to Living Frameworks}: Early benchmarks were often presented as static collections of datasets with a single canonical metric, usually accuracy. Modern frameworks like HELM \cite{HELM} and BIG-Bench \cite{BIG-Bench} are dynamic, continuously incorporating new tasks to probe an expanding range of linguistic phenomena and mitigate benchmark saturation.

\textbf{From Monolingual to Multilingual Stress Testing}: Monolingual benchmarks (GLUE \cite{GLUE}/SuperGLUE \cite{SuperGLUE}) implicitly assumed linguistic universality. Xtreme \cite{Xtreme} and MDIA \cite{MDIA} shattered this illusion by establishing typological diversity as a core robustness probe. 

\textbf{From Task-Accuracy to Multi-Dimensional Intelligence}: The LLM-as-Judge paradigm shifted evaluation from a single score like accuracy or F1 to a multi-faceted profile, assessing qualities like coherence, safety, and creativity which are previously unquantifiable at scale. Concurrently, adversarial filtering (HellaSWAG \cite{HellaSwag}, WinoGrande \cite{WinoGrande}) and synthetic data infusion (BLEURT \cite{BLEURT}) emerged as essential tools to combat dataset bias, ensuring models perform compositional reasoning rather than exploiting spurious statistical cues.

\subsubsection{Summary and Future Directions} \label{subsubsec:challenges}

The relentless evolution of benchmarks has exposed fractures in their ability to authentically measure core linguistic capabilities, revealing three critical gaps demanding urgent resolution. \textbf{Persistent cross-linguistic inequities} remain entrenched despite expanded multilingual coverage—typological biases continue to distort performance measurements, as morphosyntactic divergences between analytic and agglutinative languages manifest in systemic evaluation gaps. Benchmarks like Xtreme \cite{Xtreme} and MDIA \cite{MDIA} expose how shallow language tagging fails to capture structural phenomena like ergativity or vowel harmony, reducing linguistic diversity to mere metadata rather than a  embedded variable in assessment frameworks.

Compounding these limitations, \textbf{the self-referential trap of LLM-as-Judge methodologies} threatens to calcify evaluation into stylistic monocultures. When frontier models like GPT-4 assess conversational depth or instruction fidelity in MT-Bench \cite{MT-Bench} or BiGGenBench \cite{BiGGenBench}, they risk circularly validating their own generative patterns, privileging familiarity over authentic capability. This epistemological crisis demands adversarial auditing frameworks and ensembles of specialized, domain-tuned judges enforcing pluralism while preserving human alignment.

Meanwhile, \textbf{the specter of resource asymmetry} corrupts benchmark integrity at its foundation. HELM \cite{HELM}'s computational burden and MDIA \cite{MDIA}'s data scarcity for low-resource languages perpetuate exclusion. This violates linguistic justice—the principle that evaluation accessibility must scale with linguistic diversity. Emerging solutions like assessment with data decentralization, collective intelligence and dynamic task sampling offer pathways, but require rigorous fairness guarantees.

\subsection{Knowledge}

The capacity to store and accurately retrieve vast quantities of real-world information is a foundational pillar of modern Large Language Models (LLMs). These models function as veritable repositories of knowledge assimilated from extensive training corpora, making the quantification of this knowledge's extent and reliability a critical axis of evaluation. Consequently, benchmarks designed to probe this dimension have become a de-facto standard for gauging model progress. These evaluations typically simulate rigorous, "closed-book examinations," compelling models to rely solely on their internal, parameterized knowledge. This section presents a critical survey of the landscape of knowledge-oriented benchmarks, analyzing their methodological underpinnings, evolutionary trajectories, and the persistent challenges that shape future research. A summary of representative benchmarks is provided in Table \ref{tab:knowledge_summary}.

\begin{table}[h!]
\centering
\renewcommand{\arraystretch}{1}
\resizebox{\textwidth}{!}{
\begin{tabular}{llllllllll}
\hline
Benchmark & Focus & Language & Source & Data Type & Eval. & Indicators & Amount & Method & Citations \\ \hline
MMLU \cite{MMLU} & \begin{tabular}[t]{@{}l@{}}Comprehensive\\ knowledge\end{tabular} & Monolingual & Web & MCQA & AE & Accuracy & 15,908 & No & 4322 \\
MMLU-Pro \cite{Mmlu-pro} & \begin{tabular}[t]{@{}l@{}}Robust\\ knowledge\end{tabular} & Monolingual & Hybrid & MCQA & AE & Accuracy & 12,032 & No & 387 \\
GPQA \cite{GPQA} & \begin{tabular}[t]{@{}l@{}}Google-Proof\\ Q\&A\end{tabular} & Monolingual & Manual Design & MCQA & AE & Accuracy & 448 & No & 724 \\
SuperGPQA \cite{SuperGPQA} & \begin{tabular}[t]{@{}l@{}}Graduate-level\\ knowledge\end{tabular} & Monolingual & Hybrid & MCQA & AE & Accuracy & 26,529 & No & 17 \\
C-Eval \cite{C-Eval} & \begin{tabular}[t]{@{}l@{}}Chinese eval.\\ suite\end{tabular} & Monolingual & Web & MCQA & AE & Accuracy & 13,948 & No & 204 \\
AGIEval \cite{Agieval} & \begin{tabular}[t]{@{}l@{}}Human-centric\\ exams\end{tabular} & Bilingual & Std. Exams & Hybrid & AE & Accuracy, EM & 8,062 & No & 480 \\
GAOKAO-Bench \cite{GAOKAO-Bench} & \begin{tabular}[t]{@{}l@{}}Chinese college\\ exam\end{tabular} & Bilingual & Std. Exams & Hybrid & ME & \begin{tabular}[t]{@{}l@{}}Accuracy,\\ Scoring Rate\end{tabular} & 2,811 & No & 97 \\
KoLA \cite{KoLA} & \begin{tabular}[t]{@{}l@{}}Hierarchical\\ knowledge\end{tabular} & Monolingual & Open Datasets & Generation & AE & Custom Scores & 19 datasets & No & 138 \\
BIG-Bench \cite{BIG-Bench} & \begin{tabular}[t]{@{}l@{}}Extrapolating\\ capabilities\end{tabular} & Multilingual & Manual Design & Hybrid & ME & Multiple Metrics & 200k+ & No & 1628 \\
HELM \cite{HELM} & \begin{tabular}[t]{@{}l@{}}Holistic\\ evaluation\end{tabular} & Multilingual & Hybrid & Hybrid & AE & 7 Core Metrics & 17M+ & No & 1507 \\
M3Exam \cite{M3exam} & \begin{tabular}[t]{@{}l@{}}Multilingual/\\ modal exams\end{tabular} & Multilingual & Std. Exams & MCQA & AE & Accuracy & 12,317 & No & 141 \\
CMMMU \cite{CMMMU} & \begin{tabular}[t]{@{}l@{}}Chinese multi-\\ modal understanding\end{tabular} & Monolingual & Hybrid & Hybrid & AE & Accuracy & 12,012 & No & 14 \\ \hline
\end{tabular}
}
\caption{Summary of representative knowledge-oriented benchmarks. Evaluation methods are abbreviated as AE (Automated Evaluation), ME (Mixed Evaluation). Data sources are abbreviated as Std. Exams (Standardized Exams). The 'Method' column indicates if the paper proposed a new methodology (Yes/No).}
\label{tab:knowledge_summary}
\end{table}

\subsubsection{Evolution of Knowledge Evaluation Paradigms}

The trajectory of knowledge evaluation in LLMs mirrors the escalating capabilities of the models themselves, marked by a conceptual pivot from assessing information retrieval to probing internalized knowledge. Early paradigms often centered on open-domain question answering, such as in TriviaQA \cite{TriviaQA} and NaturalQuestions \cite{NaturalQuestions}, where models were primarily evaluated on their ability to locate answers within provided documents. 

A seminal shift occurred with the introduction of MMLU \cite{MMLU}, which established a new and influential paradigm. By presenting a massive, multi-task benchmark of multiple-choice questions across 57 diverse disciplines without external context, MMLU forced the evaluation to focus squarely on the models' parameterized knowledge. This established a rigorous standard and catalyzed an arms race in both model development and benchmark design. In response to emergent model saturation on MMLU, subsequent benchmarks have pushed the frontiers of difficulty and scope. For instance, MMLU-Pro \cite{Mmlu-pro} raised the adversarial bar by increasing the number of choices and the proportion of reasoning-intensive questions. Concurrently, benchmarks like GPQA \cite{GPQA} were designed by domain experts to be ``Google-Proof,'' directly addressing the challenge of models retrieving answers from web search rather than relying on internalized knowledge, while SuperGPQA \cite{SuperGPQA} further escalated the challenge into hundreds of highly specialized, graduate-level domains. This evolutionary arc reflects a continuous effort to create evaluations that remain challenging for even the most capable models.

\subsubsection{Methodological Landscape and Divergent Philosophies}

While sharing the common goal of knowledge assessment, these benchmarks are built upon a set of shared methodological foundations yet exhibit divergent evaluation philosophies. The predominant evaluation format is Multiple-Choice Question Answering (MCQA), a choice motivated by its scalability and amenability to objective, automated evaluation using accuracy as the primary metric. This approach, while logistically advantageous, has inherent limitations in assessing the nuances of knowledge generation and reasoning.

Beyond this common architecture, a number of distinct philosophical trajectories have emerged. \textbf{One prominent trajectory is the pursuit of human-centric alignment}, where evaluation is grounded in established human standards. Benchmarks like AGIEval \cite{Agieval} and GAOKAO-Bench \cite{GAOKAO-Bench} epitomize this approach by curating questions directly from high-stakes human examinations (e.g., college entrance and professional qualification tests). This methodology offers a more interpretable measure of a model's capabilities relative to human intellect. \textbf{Another direction focuses on achieving finer-grained analysis}. KoLA \cite{KoLA}, for example, moves beyond a single accuracy score to propose a hierarchical framework that dissects knowledge into levels of recall, understanding, and application. \textbf{A third philosophy advocates for holistic and multi-faceted evaluation}. Rather than isolating knowledge, benchmarks like HELM \cite{HELM} and BIG-Bench \cite{BIG-Bench} integrate knowledge assessment (as accuracy) into a broader suite of metrics, including robustness, fairness, and calibration, providing a more comprehensive profile of model behavior. Finally, the expansion towards \textbf{multilingual and multimodal knowledge}, exemplified by benchmarks such as the multilingual M3Exam \cite{M3exam} and the Chinese-centric multimodal benchmarks GAOKAO-MM \cite{GAOKAO-MM} and CMMMU \cite{CMMMU}, marks a critical effort to generalize evaluation beyond English-only, text-based paradigms.

\subsubsection{Summary and Future Directions}

In summary, while knowledge-oriented benchmarks have evolved to become more rigorous and diverse, they continue to face critical challenges that define the key directions for future research. \textbf{The first and most pervasive challenge is the specter of data contamination}. As models are trained on ever-expanding web-scale datasets, the probability of benchmark questions being present in the training data increases, potentially inflating scores and compromising the validity of the evaluation. This necessitates the development of dynamic or "Google-Proof" benchmarks, as seen in GPQA, as well as robust statistical methods for detecting contamination.

\textbf{A second challenge lies in the methodological limitations of closed-form evaluation}. The dominance of the MCQA format, while scalable, fails to capture a model's ability to generate coherent explanations, synthesize information, or admit uncertainty. This limitation may reward models adept at pattern matching rather than genuine comprehension. Consequently, a move towards hybrid evaluation frameworks that incorporate open-ended generation, assessed by either human experts or increasingly sophisticated LLM-as-a-judge systems \cite{MT-Bench, AlpacaEval}, is a crucial future direction.

\textbf{Finally, the issues of static evaluation and cultural bias are intertwined}. Most benchmarks represent a static snapshot of knowledge at a particular time and, often, from a predominantly Western, English-centric perspective. This not only makes them incapable of assessing a model's grasp of evolving, real-time information but also risks penalizing models with different cultural or linguistic knowledge bases. Addressing this requires a concerted effort to build more dynamic, culturally diverse, and multilingual benchmarks, following the path forged by comprehensive Chinese-language suites like CLUE \cite{CLUE} and C-Eval \cite{C-Eval}.

\subsection{Reasoning}
The ability to reason—spanning formal logic, commonsense inference, and applied problem-solving—is a cornerstone of higher intelligence. Evaluating this capability in Large Language Models (LLMs) is crucial for understanding their cognitive limits and practical potential. This section surveys a wide array of benchmarks designed to test these facets of reasoning, from structured logical puzzles to complex, real-world scenarios. A comprehensive overview of these benchmarks, categorized by reasoning type, is presented in Table \ref{tab:combined-reasoning-benchmarks-cited}.

\subsubsection{Logical Reasoning}
The domain of logical reasoning represents the most mature and densely populated area of LLM evaluation. This focus is understandable, as formal logic provides the bedrock of structured thought. The overall landscape reveals a clear developmental arc, beginning with foundational benchmarks testing discrete deductive steps (e.g., SimpleLogic~\cite{SimpleLogic}) and evolving towards assessments of highly complex, multi-step, and even programmatic reasoning (e.g., LogicPro~\cite{LogicPro}). This progression reflects the community's growing ambition, moving from asking "Can LLMs perform logical operations?" to "Can LLMs think like a reasoner?".

A primary commonality across these benchmarks is their reliance on controlled environments where logical correctness is unambiguous. As shown in Table \ref{tab:combined-reasoning-benchmarks-cited}, most datasets are either human-authored (e.g., FOLIO~\cite{FOLIO}) or synthetically generated (e.g., LogicBench~\cite{LogicBench}, ProofWriter~\cite{ProofWriter}), facilitating automated evaluation where Accuracy is the dominant metric. However, the uniqueness of these benchmarks lies in the specific facets of logic they target. We see a rich tapestry of challenges, from verifying natural language statements against first-order logic rules (LogicNLI~\cite{LogicNLI}) and solving constraint-satisfaction puzzles (ZebraLogic~\citep{ZebraLogic}, SATBench~\cite{SATBench}) to generating verifiable proofs (ProofWriter~\cite{ProofWriter}). This diversity allows for a fine-grained diagnosis of model capabilities, exposing specific failure points in their reasoning processes, such as compositional generalization (LLM\_Compose~\cite{LLM_Compose}, PrOntoQA-OOD~\cite{PrOntoQA-OOD}).

Several key trends and challenges are shaping the future of this domain. First, there is a clear push towards scalability and complexity, exemplified by datasets like LogicPro~\cite{LogicPro} with its 540,000 program-guided examples and the intricate, long-context challenges in DebateBench~\cite{DebateBench}. A second trend is the move towards programmatic and verifiable reasoning, where models generate structured outputs like code that can be executed and checked, providing more robust evaluation than simple string matching. The primary challenge remains bridging the gap between formal logic and the nuances of natural language. A further challenge is the brittleness of accuracy as a metric; future work must continue developing benchmarks that not only measure correctness but also evaluate the faithfulness and efficiency of the reasoning chain itself.

\subsubsection{Specialized and Commonsense Reasoning}
This category of benchmarks signifies a crucial expansion of the field, acknowledging that intelligence requires more than formal logic. It delves into the nuanced, often implicit, reasoning that underpins daily human cognition, such as understanding causality, leveraging commonsense knowledge, and performing mathematical calculations. The landscape here is newer and more diverse than that of pure logic, reflecting a frontier of active research united by the goal of quantifying abilities that are critical for real-world interaction.

While many of these benchmarks retain scalable automated evaluation, we see the introduction of more novel evaluation methods tailored to specific reasoning types. As detailed in Table \ref{tab:combined-reasoning-benchmarks-cited}, these include LLM-based judges to assess open-ended causal explanations (CRAB~\cite{CRAB}) and specialized metrics like the Mahalanobis distance for evaluating analogical reasoning (ANALOGICAL~\cite{ANALOGICAL}). Their uniqueness is their strength. Benchmarks like Corr2Cause~\cite{Corr2Cause} and CLadder~\cite{CLadder} pioneer the evaluation of causal inference, a critical step towards moving models from correlation to understanding. Others, like the highly-cited StrategyQA~\cite{StrategyQA} and aNLI~\cite{aNLI}, probe the implicit, multi-step, and abductive reasoning that is central to human problem-solving. Furthermore, the emergence of benchmarks for active reasoning (AR-Bench~\cite{AR-Bench}) and linguistic rule induction (IOLBENCH~\cite{iolbench}) represents a paradigm shift, moving evaluation from passive pattern recognition to active, agentic problem-solving.

\subsubsection{Applied and Contextual Reasoning}
This final category represents the crucible where all forms of reasoning are tested: the complex, noisy, and practical world of applied knowledge. These benchmarks assess an LLM's ability to deploy its skills to solve multi-faceted problems that mirror real-world tasks, serving as capstone evaluations of the entire pipeline of information retrieval, integration, reasoning, and synthesis. These are typically large-scale efforts, often with high citation counts (e.g., HotpotQA~\cite{HotpotQA}, SuperGLUE~\cite{SuperGLUE}, ARC~\cite{ARC}), marking them as flagship measures of progress in artificial intelligence.

The common thread uniting these benchmarks is their grounding in realistic, web-scale data and their focus on tasks requiring integrative reasoning. For instance, HotpotQA~\cite{HotpotQA} demands that a model locate and connect disparate pieces of evidence for Multi-hop Inferential Reasoning, while ARC~\cite{ARC} requires the application of scientific knowledge. Their uniqueness comes from the specific, complex reasoning processes they target. BIG-Bench Hard~\cite{BIG-Bench-Hard} is distinguished by its focus on Challenging Compositional Reasoning across 23 diverse tasks, while LiveBench~\cite{LiveBench} is particularly innovative for its use of live, Private user queries, creating a dynamic challenge that inherently resists data contamination.

A dominant trend in this area is the push for greater robustness and explainability. It is no longer sufficient for a model to produce the correct answer; benchmarks increasingly demand that the model "show its work" by providing supporting evidence (HotpotQA~\cite{HotpotQA}) or by succeeding on a battery of diverse and challenging tasks (SuperGLUE~\cite{SuperGLUE}). The most significant challenge facing these benchmarks is data contamination. As their Web-sourced data is public, preventing test sets from leaking into training corpora is nearly impossible. The creation of dynamic, non-public benchmarks like LiveBench~\cite{LiveBench} is a direct and necessary response. Future evaluation will likely move towards more dynamic, real-time, and interactive scenarios (TextGames~\cite{TextGames}) that test not just what a model knows, but its ability to adapt and reason in a constantly changing world.

\subsubsection{Summary and Future Directions}
The evaluation of reasoning in LLMs has evolved significantly, progressing from siloed tests of formal logic to complex, integrated assessments that mirror real-world demands. Our survey, summarized in Table \ref{tab:combined-reasoning-benchmarks-cited}, reveals a clear trajectory across three major categories. The journey begins with Logical Reasoning, where controlled, often synthetic datasets are used to probe deductive and formal abilities. It then expands into Specialized and Commonsense Reasoning, tackling more nuanced domains like causality, mathematics, and abductive inference, often requiring novel evaluation metrics beyond simple accuracy. Finally, Applied and Contextual Reasoning serves as a capstone, evaluating the synthesis of all reasoning skills on complex, multi-hop tasks drawn from web-scale data. Methodologically, this evolution is marked by a shift from human- or model-generated data towards web-crawled and now live, private data sources; a diversification of evaluation from accuracy-based automated scoring to include LLM judges and specialized metrics; and an increase in task complexity from single-step classification to interactive, multi-step generation.

Building on these trends and identified gaps, several key future directions emerge for the field:

\textbf{Embracing Dynamic and Interactive Evaluation}: The challenge of data contamination in static, web-sourced benchmarks (e.g., HotpotQA, SuperGLUE) is a critical threat to valid assessment. The future lies in dynamic benchmarks like LiveBench~\cite{LiveBench}, which use a continuous stream of new, private data. This paradigm should be expanded. Furthermore, a move towards more interactive environments, as initiated by AR-Bench~\cite{AR-Bench} and TextGames~\cite{TextGames}, is essential for evaluating agentic reasoning, where models must plan, act, and adapt based on feedback.

\textbf{Deepening the Evaluation of Reasoning Processes}: Current evaluations predominantly focus on the final output. Future benchmarks must increasingly scrutinize the reasoning process itself. This involves not just demanding a chain of thought, but developing metrics to assess its faithfulness, logical consistency, and efficiency. The verifiable, program-guided approach of LogicPro~\cite{LogicPro} is a promising step. Future work could involve causal tracing to understand which parts of a model's knowledge and context influenced its final decision, moving beyond correctness to true explainability.

\textbf{Expanding to Underexplored Reasoning Domains and Languages}: While deductive reasoning is well-covered, other critical forms of reasoning remain underexplored. The development of robust benchmarks for abductive (e.g., aNLI~\cite{aNLI}), analogical (e.g., ANALOGICAL~\cite{ANALOGICAL}), and especially causal reasoning (e.g., CLadder~\cite{CLadder}) is a pressing need. Moreover, the vast majority of reasoning benchmarks are monolingual (English). Creating multilingual and cross-lingual reasoning challenges, building on initial efforts like Multi-LogiEval~\cite{Multi-LogiEval}, is vital for ensuring that progress in AI reasoning is equitable and globally applicable.

\textbf{Integrating Reasoning with Action and Tools}: The ultimate test of reasoning is its application to achieve goals in the world. The next frontier of evaluation will require LLMs to function as agents that use tools, search for information, and interact with complex systems. This moves beyond text-based problems to scenarios where reasoning directly informs actions with tangible outcomes, representing the convergence of reasoning, planning, and agency.

\begin{table*}[h!]
\centering
\renewcommand{\arraystretch}{1}
\resizebox{\textwidth}{!}{
\begin{tabular}{llllllllll}
\hline
\textbf{Benchmark} & \textbf{Focus} & \textbf{Language} & \textbf{Source} & \textbf{Data type} & \textbf{Eval.} & \textbf{Indicators} & \textbf{Amount} & \textbf{Method} & \textbf{Citations} \\ \hline
\multicolumn{10}{l}{\textbf{\textit{Logical Reasoning}}} \\
RuleTaker~\cite{RuleTaker} & Deductive Reasoning & Monolingual & Human & Classification & AE & Accuracy & 1.2M & Yes & 375 \\
ProofWriter~\cite{ProofWriter} & Proof Generation & Monolingual & Model & Hybrid & AE & Accuracy & 1.2M+ & Yes & 275 \\
LogicNLI~\cite{LogicNLI} & First-Order Logic & Monolingual & Human & Classification & AE & \begin{tabular}[c]{@{}l@{}}Accuracy, \\ P-EM, P-AC\end{tabular} & 96,000 & No & 91 \\
Unigram-FOL~\cite{Unigram-FOL} & First-Order Logic & Monolingual & Human & Classification & AE & Accuracy & 100K & Yes & 2 \\
ReClor~\cite{ReClor} & Reading Comprehension & Monolingual & Web & MCQA & AE & Accuracy & 6,138 & Yes & 338 \\
LogiQA~\cite{LogiQA} & Reading Comprehension & Monolingual & Web & MCQA & AE & Accuracy & 8,678 & No & 316 \\
FOLIO~\cite{FOLIO} & First-Order Logic & Monolingual & Human & Generation & AE & Accuracy & 1,438 & Yes & 114 \\
P-FOLIO~\cite{P-FOLIO} & First-Order Logic & Monolingual & Human & Classification & AE & Accuracy & 19,000 & Yes & 5 \\
LogicBench~\cite{LogicBench} & Logical Patterns & Monolingual & Model & Hybrid & AE & Accuracy & 2,020 & No & 64 \\
Multi-LogiEval~\cite{Multi-LogiEval} & Multi-task Logic & Bilingual & Hybrid & Hybrid & AE & \begin{tabular}[c]{@{}l@{}}Accuracy, \\ F1\end{tabular} & 25,000 & No & 1 \\
ZebraLogic~\cite{ZebraLogic} & Matrix-based Puzzles & Monolingual & Human & Generation & AE & Cell/Puzzle Acc. & 1,000 & No & 15 \\
FineLogic~\cite{FineLogic} & Fine-grained Logic & Monolingual & Human & Classification & AE & Accuracy & 1,175 & Yes & 8 \\
LogicAsker~\cite{LogicAsker} & Atomic Logical Rules & Monolingual & Human & Classification & AE & Accuracy & 5,200 & No & 20 \\
LogicInference~\cite{LogicInference} & Long-Tailed Inference & Monolingual & Hybrid & Classification & AE & Accuracy & 9,990 & Yes & 21 \\
SimpleLogic~\cite{SimpleLogic} & Systematic Generalization & Monolingual & Human & Classification & AE & Accuracy & 7,000 & Yes & 145 \\
AutoLogi~\cite{AutoLogi} & Logic Puzzles & Bilingual & Model & LLM & Accuracy & 2,300 & No & 1 \\
SATBench~\cite{SATBench} & SAT Problems & Monolingual & Hybrid & Classification & AE & Accuracy & 2,100 & No & 1 \\
PARAT~\cite{PARAT} & SAT Solving & Monolingual & Human & Classification & AE & Accuracy & 100K+ & Yes & 1 \\
LogicPro~\cite{LogicPro} & Program-guided Logic & Monolingual & Model & Generation & LLM & Accuracy & 540K & No & 3 \\
DebateBench~\cite{DebateBench} & Long-context Debate & Monolingual & Web & Classification & AE & \begin{tabular}[c]{@{}l@{}}Position Diff., \\ Score\end{tabular} & 256 speeches & Yes & 0 \\
PrOntoQA~\cite{PrOntoQA} & "Greedy" Chain-of-Thought & Monolingual & Human & Generation & AE & Accuracy & 3 tasks & Yes & 323 \\
PrOntoQA-OOD~\cite{PrOntoQA-OOD} & Compositional Generalization & Monolingual & Web & Generation & AE & Accuracy & 1,760 & Yes & 77 \\
LLM\_Compose~\cite{LLM_Compose} & Compositional Generalization & Monolingual & Human & Generation & AE & Accuracy & Adaptive & Yes & 60 \\
\hline
\multicolumn{10}{l}{\textbf{\textit{Specialized and Commonsense Reasoning}}} \\
StrategyQA~\cite{StrategyQA} & Multi-step Strategy & Monolingual & Human & Classification & AE & Accuracy & 2,780 & Yes & 700 \\
aNLI~\cite{aNLI} & Abductive/Commonsense & Monolingual & Human & MCQA & AE & Accuracy & 169K & Yes & 800 \\
CommonGen~\cite{CommonGen} & Generative Commonsense & Monolingual & Web & Generation & AE & \begin{tabular}[c]{@{}l@{}}SPICE, \\ BLEU-4\end{tabular} & 77K & No & 600 \\
CLUTRR~\cite{CLUTRR} & Inductive Reasoning & Monolingual & Human & Classification & AE & Accuracy & 6,016 & Yes & 230 \\
Corr2Cause~\cite{Corr2Cause} & Causal Reasoning & Monolingual & Hybrid & Classification & AE & Accuracy, F1 & 1,000 & Yes & 63 \\
CRAB~\cite{CRAB} & Causal Reasoning & Monolingual & Human & Generation & LLM & Win Rate & 3,923 & Yes & 17 \\
CLadder~\cite{CLadder} & Causal Reasoning & Monolingual & Model & AE & Hybrid & Accuracy & 10K & Yes & 24 \\
MathQA~\cite{MathQA} & Mathematical Reasoning & Monolingual & Web & MCQA & AE & Accuracy & 37,298 & Yes & 706 \\
GSM-Symbolic~\cite{GSM-Symbolic} & Symbolic Math & Monolingual & Web & Generation & AE & Accuracy & 8,500 & Yes & 21 \\
Mathador-LM~\cite{Mathador-LM} & Mathematical Reasoning & Monolingual & Human & Generation & AE & Accuracy & N/A & Yes & 8 \\
MathEval~\cite{MathEval} & Mathematical Reasoning & Bilingual & Hybrid & Generation & AE & Accuracy & 64,171 & No & 23 \\
AR-Bench~\cite{AR-Bench} & Active Reasoning & Monolingual & Human & Generation & AE & \begin{tabular}[c]{@{}l@{}}Success Rate, \\ Efficiency\end{tabular} & 5,500 & Yes & 0 \\
IOLBENCH~\cite{iolbench} & Linguistic Reasoning & Multilingual & Web & Hybrid & AE & Accuracy & 1,500 & Yes & 1 \\
ANALOGICAL~\cite{ANALOGICAL} & Long-text Analogy & Monolingual & Hybrid & Classification & AE & Mahalanobis dist. & 13 datasets & Yes & 34 \\
StructTest~\cite{StructTest} & Structured Output & Monolingual & Human & Generation & AE & Rule Compliance & N/A & Yes & 2 \\
ProtoQA~\cite{ProtoQA} & Prototypical Common-sense & Monolingual & Human & Generation & AE & \begin{tabular}[c]{@{}l@{}}Exact Match,\\ Similarities,\\ Max @K\end{tabular} & 10K & Yes & 62 \\
\hline
\multicolumn{10}{l}{\textbf{\textit{Applied and Contextual Reasoning}}} \\
ARC~\cite{ARC} & Scientific Reasoning & Monolingual & Web & MCQA & AE & Accuracy & 7,787 & Yes & 1693 \\
SuperGLUE~\cite{SuperGLUE} & Broad-Spectrum Reasoning & Monolingual & Web & Hybrid & AE & \begin{tabular}[c]{@{}l@{}}Accuracy, F1, \\ EM\end{tabular} & \textasciitilde110K & No & 4000 \\
HotpotQA~\cite{HotpotQA} & Multi-hop Inferential Reasoning & Monolingual & Web & Hybrid & AE & \begin{tabular}[c]{@{}l@{}}F1, EM, \\ Ans. Acc.\end{tabular} & 112,779 & No & 4100 \\
BIG-Bench Hard~\cite{BIG-Bench-Hard} & Challenging Compositional Reasoning & Monolingual & Web & Hybrid & AE & Accuracy, etc. & 23 tasks & No & 1677 \\
SysBench~\cite{SysBench} & Algorithmic/Planning & Monolingual & Web & Hybrid & AE & Accuracy & 10 tasks & No & 11 \\
TextGames~\cite{TextGames} & Interactive Reasoning & Monolingual & Human & Hybrid & AE & Accuracy & 8 games & No & 0 \\
LiveBench~\cite{LiveBench} & Real-world Applied Reasoning & Multilingual & Private & Generation & LLM & Win Rate & 32,156 & No & 114 \\
\hline
\end{tabular}
}
\caption{
This table provides a comprehensive overview of various benchmarks used to evaluate reasoning in LLMs, categorized into three sections.
\textbf{Logical Reasoning:} This section includes benchmarks that specifically target deductive, first-order logic, and other formal reasoning abilities. 
\textbf{Specialized and Commonsense Reasoning:} This category covers benchmarks that evaluate reasoning in specialized domains like mathematics and broader commonsense understanding. 
\textbf{Applied and Contextual Reasoning:} These benchmarks assess how well LLMs can apply their reasoning skills to complex, multi-step tasks that often mirror real-world scenarios. 
\newline 
\textit{Abbreviations: AE: Automated Evaluation; LLM: LLM-based Judge; MCQA: Multiple Choice Question Answering; EM: Exact Match; P-EM: Probability-based Exact Match; P-AC: Probability-based Accuracy. The 'Method' column indicates if the paper proposed a new methodology (Yes/No).}
}
\label{tab:combined-reasoning-benchmarks-cited}
\end{table*}

\section{Domain-Specific Benchmarks}
\subsection{Natural Sciences}

Shifting the evaluation perspective from general capabilities to specialized domains is a critical step in testing the boundaries of Large Language Models (LLMs). As one of the most logically rigorous and structurally organized areas of human knowledge, the natural sciences present a significant challenge to an LLM's knowledge base and reasoning abilities.
This field, which spans core disciplines like Mathematics, Physics, Chemistry, and Biology, shares a common set of features. Success in this area not only requires a model to have good general-purpose abilities, but also demands strong capacities for abstract reasoning, symbolic manipulation, and following complex causal chains. 
For example, a physics problem might require the application of a specific mathematical theorem, while a model must be able to refuse to answer a chemistry question about how to make explosives.

As summarized in Table~\ref{tab:natural_sciences_benchmarks}, this section will review and analyze these representative domain-specific benchmarks, discussing their design philosophies, evaluation dimensions, and the common challenges they face.
To systematically examine the performance of LLMs in different branches of the natural sciences, existing evaluation benchmarks are typically categorized by discipline, each focusing on the unique challenges of that specific field.

\begin{table}[h!]
\centering
\renewcommand{\arraystretch}{1}
\resizebox{\textwidth}{!}{
\begin{tabular}{llllllllll}
\hline
Benchmark & Focus & Language & Source & Data Type & Eval. & Indicators & Amount & Method & Citations \\ \hline
\multicolumn{10}{l}{\textbf{\textit{Mathematics}}} \\
GSM8K \cite{GSM8K} & Grade School Math & Monolingual & Manual Design & Generation & AE & Accuracy  & 85k  & Yes & 3068 \\
MATH \cite{MATH} & Competition Math & Monolingual & Hybrid & MCQA & AE & Accuracy & 12.5k & No & 1719 \\ 
U-MATH \cite{u-math}& Undergraduate Math & Monolingual & Std. Exams & Hybrid & LLM & \begin{tabular}[t]{@{}l@{}}Accuracy, \\PPV, TPR, etc.\end{tabular} & 1.1k & Yes & 12 \\ 
Omni-MATH \citep{Omni-MATH} & Olympiad Math & Monolingual & Web & Generation & AE & Accuracy & 4.4k & No & 93 \\
MiniF2F \cite{minif2f} & Formal Theorem & Monolingual & Open Datasets & Generation & AE & Pass rate & 488 & No & 208 \\
FrontierMath \cite{frontiermath} & Advanced Math & Monolingual & Manual Design & Generation & AE & Accuracy & 300 & Yes & 62 \\
MATH-P \cite{math-p} & Perturbed Math & Monolingual & Open Datasets & Generation & AE & Accuracy & 279 & Yes & 20 \\
ASyMOB \cite{asymob} & Symbolic Math & Monolingual & Hybrid & Generation & AE & Accuracy & 17k & Yes & 0 \\ 
MathBench \cite{mathbench} & Multi-difficulty Math & Bilingual & Hybrid & Hybrid & ME & Accuracy,CE & 3.7k & Yes & 80 \\
Hardmath \cite{hardmath} & Graduate Math & Monolingual & Auto Design & Generation & ME & Accuracy & 1.4k & Yes & 14 \\
Lila \cite{lila} & Multi-domain Math & Monolingual & Open Datasets & Hybrid & AE & F1 & 134k & Yes & 143 \\ 
HARP \cite{harp} & Olympiad Math & Monolingual & Web & Hybrid & AE & Accuracy & 5.4k & No & 9 \\ 
Mathador-LM \cite{Mathador-LM} & Math Game & Monolingual & Auto Design & Generation & AE & Accuracy & 1k & Yes & 8 \\ \hline
\multicolumn{10}{l}{\textbf{\textit{Physics}}} \\
PHYSICS  \cite{physics}    & Undergrad. Physics & Monolingual & Manual Design & Generation & AE    & Accuracy    & 1.3k   & No          & 0         \\
UGPhysics \cite{ugphysics}    & Undergrad. Physics & Bilingual   & Web           & Hybrid     & ME    & Accuracy    & 11k  & Yes         & 5         \\
PhysReason  \cite{physreason} & Multimodal Physics & Monolingual & Web           & Generation & AE    & Accuracy    & 1.2k   & Yes         & 9         \\
PhysicsArena \cite{physicsarena}& Multimodal Physics & Monolingual & Web           & Generation & ME    & Accuracy    & 5.1k   & Yes         & 0         \\
PhysUniBench \cite{physunibench}& Vision-Essential & Bilingual   & Std. Exams    & Hybrid     & ME    & Accuracy    & 3.3k   & No          & 0         \\
SeePhys \cite{seephys}       & Vision-Essential & Bilingual   & Std. Exams    & Generation & ME    & Accuracy    & 2k   & No          & 1     \\
FEABench \cite{FEABench}& Eng. Simulation & Monolingual & Web & Generation & AE & Resolved Ratio & 1.4k & Yes & 2 \\ 
TPBench \cite{tpbench} & Theoretical Physics & Monolingual & Hybrid & Generation & ME & Accuracy & 57 & Yes & 9 \\ \hline

\multicolumn{10}{l}{\textbf{\textit{Chemistry}}} \\
ChemEval  \cite{chemeval}   & Foundational Chem. & Monolingual & Hybrid & Hybrid & ME & Accuracy, F1 & 1.5k & No & 7 \\
ChemistryQA \cite{chemistryqa}  & Literature-based & Monolingual & Web & Generation & AE & Acc, Precision & 4.4k & No & 3 \\
ScholarChemQA \cite{scholarchemqa}& Literature-based & Monolingual & Web & MCQA & AE & Accuracy, F1 & 40k & No & 7 \\
MoleculeQA \cite{moleculeqa}    & Molecular Prop. & Monolingual & Open Datasets & MCQA & AE & Accuracy & 61.5k & No & 9 \\
ChemSafetyBench \cite{chemsafetybench}& Chemical Safety & Monolingual & Web & Generation & LLM & Acc, Safety & 30k+ & No & 2 \\ 
ChemLLMBench \cite{chemllmbench}& Molecular Prop. & Monolingual & Open Datasets & Hybrid & AE & \begin{tabular}[t]{@{}l@{}}Accuracy,F1\\BLEU,ROUGE\\Validity,Exact Match\end{tabular} & 100k & No & 216 \\ 
Alchemy \cite{alchemy}& Molecular Prop. & Monolingual & Open Datasets & Generation & AE & MAE & 110k & No & 100 \\ \hline
\multicolumn{10}{l}{\textbf{\textit{Biology}}} \\
PubMedQA \cite{pubmedqa} & Biomedical QA & Monolingual & Web & MCQA & AE & Accuracy, F1 & 274k & Yes & 984 \\
BioMaze  \cite{biomaze}  & Pathway Reasoning & Monolingual & Web & Hybrid & ME & Accuracy & 5.1k & No & 0 \\
SciAssess \cite{sciassess}& Paper Analysis & Monolingual & Open Datasets & Hybrid & AE & \begin{tabular}[t]{@{}l@{}}Acc, Recall, F1,\\ Mol. Similarity\end{tabular} & 6.9k & No & 29 \\ 
BioPreDyn-bench \cite{biopredyn}& Biological modeling & Monolingual & Manual Design & Generation & AE & NRMSE & 6 & No & 99 \\ 
BixBench \cite{BixBench}& Computational Biology & Monolingual & Manual Design & Generation & LLM & Accuracy & 296 & Yes & 8 \\
LAB-Bench \cite{labbench}& Biology research & Monolingual & Hybrid & MCQA & AE & \begin{tabular}[t]{@{}l@{}}Accuracy\\Precision,Coverage\end{tabular} & 2.4k & Yes & 54\\
AutoBio \cite{autobio}& Biological experiment & Monolingual & Manual Design & Generation & AE & Accuracy & 100 & Yes & 2\\
BioProBench \cite{bioprobench} & Biological experiment & Monolingual & Open Datasets & Hybrid & AE & \begin{tabular}[t]{@{}l@{}}Acc,F1,Recall\\BLEU,METEOR\\ROUGE-L\\EM \end{tabular}& 556k & Yes & 1\\ \hline
\multicolumn{10}{l}{\textbf{\textit{Cross-Disciplinary}}} \\
\multirow{3}{*}{JEEBench \cite{JEEBench}} & Math & \multirow{3}{*}{Monolingual} & \multirow{3}{*}{Web} & \multirow{3}{*}{Hybrid} & \multirow{3}{*}{ME} & \multirow{3}{*}{Accuracy} & \multirow{3}{*}{515} & \multirow{3}{*}{No} & \multirow{3}{*}{63} \\ 
& Physics & & & & & & & & \\
& Chemistry & & & & & & & & \\ \hline
\multirow{4}{*}{SciBench \cite{scibench}} & Math & \multirow{4}{*}{Monolingual} & \multirow{4}{*}{Web} & \multirow{4}{*}{Generation} & \multirow{4}{*}{ME} & \multirow{4}{*}{Accuracy} & \multirow{4}{*}{972} & \multirow{4}{*}{Yes} & \multirow{4}{*}{162} \\
& Physics & & & & & & & & \\
& Chemistry & & & & & & & & \\ 
& CS & & & & & & & & \\ \hline
TheoremQA \cite{theoremqa} & Multi-domain & Monolingual & Manual Design & Hybrid & AE & Accuracy & 800 & No & 150 \\
OpenBookQA \cite{OpenBookQA} & Sci. Commonsense & Monolingual & Manual Design & MCQA & AE & Accuracy & 5.9k & No & 1171 \\
GPQA \cite{GPQA} & \begin{tabular}[t]{@{}l@{}}Physics, Chem.,\\ Biology\end{tabular} & Monolingual & Manual Design & MCQA & AE & Accuracy & 1.1k & No & 724 \\ 
WMDP \cite{wmdp}& Biology,Chem.,CS & Monolingual & Web & MCQA & AE & Accuracy & 3.6k & Yes & 226 \\ 
Curie \cite{curie} & \begin{tabular}[t]{@{}l@{}}Physics,Chem.,\\Biology\end{tabular} & Monolingual & Web & Generation & ME & \begin{tabular}[t]{@{}l@{}}ROUGE-L, \\BERTScore F1, \\IoU, IDr, \\LMScore, LLMSim\end{tabular} & 580 & Yes & 2 \\\hline
\end{tabular}
} 
\caption{Summary of representative benchmarks in the Natural Sciences. Data sources are abbreviated as Std. Exams (Standardized Exams). The 'Method' column indicates if the paper proposed a new methodology (Yes/No).}
\label{tab:natural_sciences_benchmarks}
\end{table}

\subsubsection{Mathematics}
Mathematics is the language of the natural science, evaluating a model's performance in this area is fundamental to measuring its abstract and logical reasoning capabilities. The evaluation of a model's mathematical capabilities is similar to human examinations, primarily utilizing multiple-choice questions and open-ended problems.

With the rapid advancement of Large Language Model (LLM) capabilities, the difficulty of mathematical evaluation benchmarks has increased.GSM8K \cite{GSM8K}, which \textbf{focus on grade school-level} word problems requiring models to perform multi-step arithmetic operations. To increase the difficulty, MATH \cite{MATH} and JEEBench \cite{JEEBench} collect problems from \textbf{high school and university entrance competitions}, covering more complex topics in algebra, geometry, and other fields. As model capabilities have grown, the difficulty of benchmarks has continued to rise, leading to several benchmarks aimed at higher levels. U-MATH \cite{u-math} cover \textbf{undergraduate-level} mathematics problems; Omni-MATH \cite{Omni-MATH} and MiniF2F \cite{minif2f} focus on \textbf{Olympiad-level} problems and formal theorem proving;  while FrontierMath~\cite{frontiermath}, designed by top mathematicians, aims to evaluate a model's ability to solve \textbf{cutting-edge advanced mathematics problems}, representing the current peak of difficulty in mathematical reasoning evaluation.

Mathematical evaluation benchmarks have a significant portion consists open-ended questions, with accuracy or pass rate serving as the core metric.This paradigm introduces the outcome-based problem that \textbf{open-ended problems without partial credit}, a right reasoning process can receive zero points due to a minor calculation error. To address this issue, new evaluation paradigms have been proposed. MATH-P \cite{math-p} tests model robustness and generalization by applying difficult perturbations to problems; ASyMOB \cite{asymob} focuses on university-level symbolic mathematical operations to assess a model's symbolic manipulation skills; and U-MATH \cite{u-math} introduces \textbf{LLM-as-a-Judge} evaluation method for more nuanced assessment.

\subsubsection{Physics}
As a bridge connecting the abstract world of mathematics with the physical world, physics presents unique demands on the reasoning capabilities of Large Language Models (LLMs). Physics problems require not only mathematical computation but also a profound conceptual understanding, the ability to ground abstract problems in physical laws. 

Early \textbf{comprehensive scientific benchmarks} laid the foundation for physical evaluation, and were subsequently followed by the emergence of more specialized and in-depth physics benchmarks. SciBench \cite{scibench} is an early comprehensive science benchmark at the university level, it covers chemistry, physics, and mathematics three disciplines. It is designed to \textbf{test a model's capabilities in multi-step reasoning, understanding scientific concepts, knowledge retrieval, and complex numerical calculations.} The majority of other benchmarks follow this same evaluation paradigm, like JEEEBench \cite{JEEBench}. PHYSICS \cite{physics} and UGPhysics \cite{ugphysics} have built English and Chinese-English undergraduate-level physics problem sets. Respectively,  UGPhysics \cite{ugphysics} specifically designed to prevent data leakage and it was found that LLMs specialized for mathematics is not always outperform other models, while PhysReason \cite{physreason} and PhysicsArena \cite{physicsarena} have introduced a large number of multimodal problems that require analysis of diagrams and charts. 

\textbf{Diagrams are often function as an indispensable part of the problem itself in physics.} Therefore, multimodal questions are an essential component of physics evaluation benchmarks.PhysUniBench \cite{physunibench} equips each problem with a corresponding diagram, while SeePhys \cite{seephys} designs the majority of its problems to be vision-essential. Furthermore, PhysicsArena \cite{physicsarena} introduces a fine-grained multimodal evaluation paradigm based on the physics problem-solving process, includes variable identification, physical process modeling, and reasoning-based solving 3 stages, moving beyond a single answer judgment.

\textbf{Trends and challenges} Physics reasoning is fundamentally more than solving "mathematical word problems," as it requires a unique, comprehensive set of capabilities that transcend simple mathematics, including \textbf{Conceptual Grounding, Multimodal Interpretation, and Process Formulation}. Like the evaluation of mathematical problems, a simple outcome-based approach cannot evaluate a model's capabilities in physics comprehensively. Such as the MARJ framework within UGPhysics and methods like LLM-as-a-Judge are trying to overcome this limitation. However, a model's physics capabilities must be grounded in practical applications, requiring it to move beyond rote problem-solving to the construction of accurate physical models for real-world scenarios. Like the way forged by FEABench \cite{FEABench}, the evaluation criteria shift from correctness to the ability to construct a valid physical model and derive verifiable results from simulation software.

\subsubsection{Chemistry}
Chemical evaluation benchmark not only focus on traditional problem-solving abilities but also extends to the critical areas of factual accuracy, the comprehension of literature, and the model's understanding of safety.ChemEval \cite{chemeval} have established multi-level evaluation systems to evaluate model's foundational knowledge. ChemistryQA \cite{chemistryqa} and ScholarChemQA \cite{scholarchemqa} extract questions from chemical literature and papers to assess a model's understanding of scientific texts. Regarding the evaluation of LLMs in subfields of chemistry, MoleculeQA \cite{moleculeqa} builds a large-scale dataset specifically to evaluate the capability of model's regarding molecular structure, properties, and more. ChemSafetyBench \cite{chemsafetybench} as a pioneering work in chemical safety area, construct a massive test set of over 30,000 samples to systematically evaluate a model's safety and responsibility when handling potentially hazardous chemical knowledge. 

Chemistry benchmarks uniquely place non-technical and social dimensions—accuracy and safety—at the core of their evaluation, to an extent that surpasses benchmarks in mathematics and physics. While the focus in mathematics and physics remains on the correctness of solutions and the logical consistency of the reasoning process, chemistry domain has build benchmarks like MoleculeQA \cite{moleculeqa}, which is entirely focus on verifying factual accuracy, and ChemSafetyBench \cite{chemsafetybench}, designed to assess safety and ethical risks.This divergence in focus origin from the different real-world implications of these disciplines: an incorrect mathematical answer is merely an error, but an incorrect chemical statement can be actively dangerous. When an LLM generates false information about molecular properties or provides synthesis methods for hazardous substances, it can directly lead to real-world harm. This indicates that as LLM evaluation engages with disciplines more closely tied to the real world, the standard of a "good" model is expanding beyond mere problem-solving ability to encompass a broader range of capabilities, including reliability, trustworthiness, and ethical alignment.

\subsubsection{Biology}
Biology evaluation benchmark primarily focus on the comprehension of relevant scientific literature.Building on classic biomedical question-answering benchmarks like PubMedQA \cite{pubmedqa}, new benchmarks are expanding into more specialized and in-depth reasoning tasks. BioMaze \cite{biomaze} focuses on reasoning about biological pathways, requiring models to understand and predict the downstream effects that arise when a biological system is subjected to interventions, such as gene mutations, viral infections, or drug treatments. SciAssess \cite{sciassess} is dedicated to evaluating a model's ability to analyze biology literature in real scientific research scenarios, with tasks ranging from basic knowledge to advanced analytical reasoning. AutoBio \cite{autobio} and BioProBench \cite{bioprobench} introduce a new paradigm for assessing biological competence by conducting biological experiments or evaluating experimental protocols to test the LLM's understanding of experimental standards.

The uniqueness of biology lies in its vast, fragmented, and often incomplete knowledge graph. While complexity exists in other scientific fields, in biology, it is concentrated in the complex, multi-step biological pathways composed of genes, proteins, and metabolites. Early benchmarks primarily tested text comprehension. However, BioMaze \cite{biomaze} highlights that true biological reasoning involves understanding complex networks, where a minor perturbation can trigger a cascade of non-linear biological chain reactions. It introduced the PATHSEEKER agent, which integrates an LLM with structured navigation of a biological knowledge graph, propelling "Graph-Augmented LLMs" as a highly promising direction in the field of biology.

\subsubsection{Cross-Disciplinary and General Scientific Abilities}

True scientific research is often interdisciplinary, so a series of benchmarks have been designed to evaluate a model's comprehensive scientific capability.\textbf{Comprehensive Problem Solving} benchmarks like JEEBench \cite{JEEBench} (Physics, Chemistry, Mathematics), SciBench \cite{scibench}(Physics, Chemistry, Mathematics, Computer Science), and GPQA \cite{GPQA} (Biology, Physics, Chemistry) assess a model's overall problem-solving skills through difficult questions spanning multiple disciplines. GPQA \cite{GPQA} in particular, is authored by domain experts and is designed to be "Google-Proof" to effectively test models.\textbf{Higher-Order Reasoning and Tool Use} benchmarks like TheoremQA \cite{theoremqa} requires models to apply theorems from disciplines like mathematics and physics to solve problems across fields. LLM-SRBench  \cite{LLM-SRBench}focuses on discovering equations from data. In natural science domain is also gradually beginning to test their \textbf{ability to use tools.} SCITOOLBENCH \cite{sciagent} provides a series of API tools that models must call to solve complex scientific calculations and reasoning tasks, has taken a key step in this direction. In addition to specialized domains, benchmarks like OpenBookQA \cite{OpenBookQA}, SciEval \cite{scieval}, and SciKnowEval \cite{sciknoweval} assess a model's \textbf{common sense science knowledge}, multi-level scientific knowledge, and research capabilities from an overall perspective, while CURIE \cite{curie} focuses on evaluating a model's understanding in long scientific literature text. Natural sciences evaluation benchmarks are gradually expanding to encompass other disciplines. GEO-Bench \cite{geo-bench} uses Earth monitoring data to evaluate pre-trained models in processing geospatial data.

Comprehensive scientific evaluation benchmarks are evolving from testing a model's static knowledge reserve to measuring its dynamic, process-oriented application capabilities. As the development of general models increasingly orients towards creating usable "research assistants," evaluation benchmarks may also transcend the boundaries of "problem-solving." It is vital to build interactive environments capable of assessing scientific research abilities. That these benchmarks will not only provide a more holistic measure of a model's utility but also play a crucial role in breakthroughs for artificial general intelligence in the natural science domain.

\subsubsection{Summary and Future Directions}
Evaluation benchmarks across the natural sciences and other domains face similar critical challenges. \textbf{A primary concern is data contamination}. if evaluation data is included in model's training set, the assessment becomes an "open-book exam" that cannot truly measure the model's reasoning capabilities. 

\textbf{Furthermore, the reliability of evaluation methods} is under scrutiny. Traditional methods based solely on final answers are often deemed insufficient, while some paradigms like "LLM-as-a-Judge" are also being questioned for their robustness. Indeed, some reviewers have noted the self-contradiction within LLM-generated evaluation reports, casting further doubt on the reliability of these automated assessment methods. 

\textbf{Evaluating the generalization capabilities of model} is a core challenge for benchmarks, which must determine whether a model has truly understood the knowledge or has merely memorized solution templates for similar problems. Perturbation-based benchmarks, such as MATH-Perturb \cite{math-p} and ASyMOB \cite{asymob}, provide a potential approach. By making some modifications to problems, they reveal that even SOTA models often rely on "shortcut learning" rather than generalizable reasoning.

To address the challenges, several new evaluation paradigms are being explored. In the domain of the natural sciences, the role of an LLM is shifting from a simple knowledge retrieval tool to a research assistant. Therefore, evaluation benchmarks are \textbf{transitioning from assessing LLMs as "knowledge bases" to evaluating their capabilities as agents.} Following the path forged by benchmarks like SciAgent \cite{sciagent}, FEABench \cite{FEABench}, and BioMaze \cite{biomaze}, the focus now is on assessing an LLM's ability to use provided tools to approach a goal.

\textbf{A move towards Holistic, Multi-faceted Frameworks.} Evaluation is shifting from a single score to a comprehensive assessment of a model's overall capabilities. Benchmarks such as ChemEval \cite{chemeval}, SciKnowEval \cite{sciknoweval}, and PhysicsArena \cite{physicsarena} build multi-dimensional or multi-stage frameworks to fine-grained evaluate model capabilities, thereby providing more actionable guidance for model improvement.

\subsection{Humanities \& Social Sciences}

Beyond the rational evaluation of large language models (LLMs) in the natural sciences, their anthropomorphic conversational traits enable more natural and effective communication with humans, enhancing interactive applications. Social sciences, as one of the most human-centered fields, play a crucial role in this context. A key question is whether LLMs can effectively address real-world challenges in areas such as \textbf{Law}, \textbf{Intellectual Property (IP)}, \textbf{Education}, \textbf{Psychology}, and \textbf{Finance}. For example, can LLMs comprehend human emotions well enough to provide meaningful emotional support? Can they reliably retain and apply legal knowledge to offer sound legal advice? This section focuses on the human-centered capabilities of LLMs within social science domains. It reviews and analyzes relevant benchmarks that investigate these aspects, examining their task design principles, data construction methods, and evaluation strategies — including those involving subjective judgments.

All these humanities and social sciences domains are highly applicable in real-world scenarios. One of the biggest challenges is determining how to evaluate an LLM’s knowledge within these domains, which involves defining appropriate tasks, constructing relevant datasets, and selecting suitable evaluation methods. These three key aspects are precisely what existing domain-specific benchmarks focus on and claim to address. In this section, we will follow the structure of these three aspects—task definition, dataset construction, and evaluation methods—to present the content for each domain, including Law, Intellectual Property (IP), Education, Psychology, and Finance. We provide detailed information on the representative benchmarks in the humanities and social sciences discussed in each subsection below, as summarized in Table~\ref{tab:4.2}.

\begin{table*}[!h]
\centering
\renewcommand{\arraystretch}{1}
\resizebox{\textwidth}{!}{
\begin{tabular}{llllllllll}
\hline
Benchmark & Focus & Language & Source & Data Type & Eval. & Indicators & Amount & Method & Citations \\ 
\hline
\multicolumn{10}{l}{\textbf{\textit{Law}}} \\
LegalBench~\cite{Legalbench} & Law & Monolingual & Hybrid & Hybrid & ME & \begin{tabular}[t]{@{}l@{}}Exact Match, F1-Score\\Correct, Analysis\end{tabular} & 91,206 & No & 162 \\
LBOX OPEN~\cite{LBOXOPEN}&Law&Monolingual&Open Datasets&Hybrid&AE&Exact Match, F1-Score&147K&Yes&51\\
LawBench~\cite{LawBench} & Law & Monolingual & Hybrid & Hybrid & AE & \begin{tabular}[t]{@{}l@{}}Accuracy, ROUGE\\ F-Score, nLog-distance\end{tabular} & -- & No & 133 \\
LAiW~\cite{LAiW} & Law & Monolingual & Open Datasets & Hybrid & ME & \begin{tabular}[t]{@{}l@{}}Accuracy, Miss Rate, F1-Score\\Entity-Acc, ROUGE, Win Rate\end{tabular} & 11,605 & No & 39 \\
LexEval~\cite{LexEval} & Law & Monolingual & Hybrid & Hybrid & AE & Accuracy, ROUGE & 14,150 & No & 24\\
CiteLaw~\cite{CiteLaw}&Law&Monolingual&Hybrid&Generation&AE&MAUVE, ROUGE, Citation Quality&1,000&No&3\\
CaseGen~\cite{CaseGen}&Law&Monolingual&Open Datasets&Generation&ME&\begin{tabular}[t]{@{}l@{}}ROUGE, BertScore, \\ LLM Judge\end{tabular}&2,000&No&3\\
\hline
\multicolumn{10}{l}{\textbf{\textit{Intellectual Property (IP)}}} \\
PatentEval~\cite{PatentEval} & IP & Monolingual & Open Datasets & Generation & ME & \begin{tabular}[t]{@{}l@{}}SemSim, N-grams Coverage\\FactGraph, QAFactEval, EntityGrid\end{tabular} & 400 & No & 4 \\
MoZIP~\cite{MoZIP}&IP&Multilingual&Web&Hybrid&AE&Accuracy&3,121&Yes&8\\
IPEval~\cite{IPEval}&IP&Bilingual&Std. Exams&MCQA&AE&Accuracy&2,657&No&3\\
D2P~\cite{D2P}&IP&English&Open Datasets&Generation&ME&ROUGE, BLEU, BERTScore&1,933&Yes&12\\
IPBench~\cite{IPBench} & IP & Bilingual & Hybrid & Hybrid & ME & \begin{tabular}[t]{@{}l@{}}Accuracy,Exact-Match,LLM Score\\BLEU, ROUGE, BertScore\end{tabular} & 10,374 & No & 0 \\
\hline
\multicolumn{10}{l}{\textbf{\textit{Education}}} \\
E-Eval~\cite{E-EVAL}&Education&Monolingual&Web&MCQA&AE&Accuracy&4,352&No&9\\
EduBench~\cite{EduBench} & Education & Bilingual & Model & Generation & ME & \begin{tabular}[t]{@{}l@{}}Scenario Adaptation Criteria, Factual\\Reasoning Accuracy Criteria\\Pedagogical Application Criteria\end{tabular} & 4,019 & No & 0 \\
\hline
\multicolumn{10}{l}{\textbf{\textit{Psychology}}} \\
CPsyExam~\cite{CPsyExam}&Psychology&Monolingual&Web&MCQA&AE&Accuracy&22,400&No&0\\
CPsyCoun~\cite{CPsyCoun} & Psychology & Monolingual & Web & Generation & AE & \begin{tabular}[t]{@{}l@{}}Comprehensiveness, Professionalism\\Authenticity, Safety\end{tabular} & 4,700 & Yes & 32 \\
Psycollm~\cite{Psycollm}&Psychology&Monolingual&Std. Exams&Hybrid&AE&\begin{tabular}[t]{@{}l@{}}ROUGE, BLEU, BERTScore\\Standard Accuracy, Elastic Accuracy\end{tabular}&3,863 &Yes&27\\
Psychometrics Benchmark~\cite{li2024quantifying}&Psychology&Monolingual&Hybrid&Hybrid&AE&\begin{tabular}[t]{@{}l@{}} Standard Deviation, Match Rate\\Cohen's kappa coefficient, Agreement Rate\end{tabular}&3,545&No&43\\
PsychoBench~\cite{PsychoBench}&Psychology&Monolingual&Std. Exams&Hybrid&AE&Mean and Standard Deviation of the Scale&512&No&38\\
\hline
\multicolumn{10}{l}{\textbf{\textit{Finance}}} \\
BBT-CFLEB~\cite{BBT-CFLEB}&Finance&Monolingual&Web&Hybrid&AE&Rouge, F1, Accuracy&220M&Yes&86\\
FLARE~\cite{FLARE}&Finance&Monolingual&Open Datasets&Hybrid&AE&Accuracy, F1, Exact-Match&136K&Yes&202\\
FinEval~\cite{Fineval}&Finance&Monolingual&Web&Hybrid&AE&Accuracy, ROUGE&8,351&No&51\\
\hline
\end{tabular}
}
\caption{Summary of representative benchmarks in the humanities and social sciences. Data sources are abbreviated as Std. Exams (Standardized Exams). Evaluation methods (listed under the Eval. column) are abbreviated as AE (Automated Evaluation) and ME (Mixed Evaluation). Abbreviations used in the Data Type column include MCQA (Multiple-Choice Question Answering). The 'Method' column indicates if the paper proposed a new methodology (Yes/No).}
\label{tab:4.2}
\end{table*}

\subsubsection{Law}
\paragraph{Legal Task Taxonomy.}The most practical application of the legal domain in the real world is to provide legal support for clients. This requires LLMs to memorize concise statutes, apply them appropriately, and respond with clear and logical reasoning. Existing benchmarks adopt different philosophical approaches to task taxonomy in the legal domain, dividing real-world application tasks into distinct levels of cognition. Bloom’s Taxonomy~\cite{Bloom} is a commonly used framework for categorizing the cognitive capabilities of LLMs, with its latest version comprising the following levels: \textbf{Remember}, \textbf{Understand}, \textbf{Apply}, \textbf{Analyze}, \textbf{Evaluate}, and \textbf{Create}. LawBench~\cite{LawBench} proposed a legal task taxonomy based on Bloom’s Taxonomy for the domain of Chinese judicial jurisdiction, classifying legal-related abilities into three cognitive levels: \textbf{(a) legal knowledge memorization}, \textbf{(b) legal knowledge understanding}, and \textbf{(c) legal knowledge application}. These three cognitive levels are a direct adaptation and summarization of Bloom’s Taxonomy, and are further divided into 20 fine-grained tasks, including article and knowledge recall, document-level element recognition and information processing, as well as legal reasoning involving penalties based on real-world cases.

LexEval~\cite{LexEval}, also a benchmark for the Chinese judicial domain and built upon Bloom’s Taxonomy, proposes a Legal Cognitive Ability Taxonomy (LexAbility Taxonomy), which is similar to that of LawBench but more fine-grained, comprising six dimensions: \textbf{Memorization}, \textbf{Understanding}, \textbf{Logical Inference}, \textbf{Discrimination}, \textbf{Generation}, and \textbf{Ethics}. LexEval is more fine-grained than LawBench, comprising 23 tasks spanning six cognitive levels. Beyond what LawBench covers, it also incorporates key aspects of the legal domain, such as the evolution of law and ethical considerations, particularly in relation to bias, discrimination, and privacy. This helps evaluate whether LLMs effectively capture the nature of this specific legal domain framework for their human-like chat support, without introducing intrinsic bias. Beyond Bloom’s Taxonomy, LAiW~\cite{LAiW}, based on legal domain practice and also focused on the Chinese judicial jurisdiction, proposes a syllogism-based taxonomy reflecting the thinking process of legal experts and legal practice, classifying tasks into three levels from easy to difficult: \textbf{basic information retrieval}, \textbf{legal foundation inference}, and \textbf{complex legal application}. This benchmark focuses on concrete, real-world legal practice. Beyond legal reasoning and case understanding, it introduces element recognition and Named Entity Recognition (NER) tasks, specifically tailored for legal domain retrieval.

In addition to legal benchmarks specifically designed for the Chinese judicial domain, there are also benchmarks developed for other languages and legal systems. LBOX OPEN~\cite{LBOXOPEN} is the first large-scale legal benchmark specifically designed for the Korean judicial domain. Unlike cognition-oriented evaluations, it focuses on legal case analysis, including two tasks for case name and statute classification, two tasks for legal judgment prediction, and one for case summarization. To better understand the legal reasoning capabilities of LLMs, LegalBench~\cite{Legalbench} focuses primarily on legal reasoning and comprises 162 tasks, including \textbf{issue spotting}, \textbf{rule recall}, \textbf{rule application and conclusion}, \textbf{statutory interpretation}, and \textbf{rhetorical understanding}. Each of these five task types includes more scene-specific subtasks with substantial real-world application value—such as applying diversity jurisdiction tests based on information about the plaintiff, defendant, and the amount in controversy for different claims, which requires both arithmetic and logical reasoning.

All the legal benchmarks above evaluate the comprehensive legal capabilities of LLMs; however, there are also benchmarks that focus on specific legal abilities within particular scenarios. As a result, their task taxonomies are more fine-grained and concrete. CiteLaw~\cite{CiteLaw} aims to evaluate LLMs whether could produce legally sound responses with appropriate citations. Legal case documents play a critical role in the legal domain. CaseGen~\cite{CaseGen} proposes an automated legal document generation task taxonomy within the Chinese judicial domain, including tasks such as \textbf{drafting defense statements}, \textbf{writing trial facts}, \textbf{composing legal reasoning}, and \textbf{generating judgment results}. These tasks require LLMs not only to accurately recall and apply legal knowledge, but also to process the internal logic within the concrete context of each case.

Different judicial domains exhibit significant differences in legal practice. As the benchmarks mentioned above are monolingual, they typically correspond to the legal systems of a single country. There is a lack of multilingual legal benchmarks for LLMs, highlighting the need to design a cross-jurisdictional evaluation task taxonomy—such as tasks focused on legal comparison across countries and recognition of cross-system differences—which could serve as a promising direction for future research in advancing legal AI.

\paragraph{Legal Dataset Source and Construction.}Real-world legal documents and cases are the most important data sources for these benchmarks. LBOX OPEN~\cite{LBOXOPEN} compiles its corpus from Korea’s first- and second-instance trials as well as Supreme Court precedents.However, most benchmarks such as LawBench~\cite{LawBench}, LegalBench~\cite{Legalbench}, and LexEval~\cite{LexEval} rely on existing datasets derived from legal competitions like CAIL and LAIC, or on publicly available legal corpora—most of which were originally created for non-LLM evaluation settings. Considerable effort has been made to recompile and adapt these existing datasets. For example, LegalBench~\cite{Legalbench} restructures the CUAD dataset~\cite{CUAD} to formulate a binary classification task for each type of contractual clause. The last approach to dataset construction in these legal benchmarks involves substantial human effort, often including the participation of legal professionals to manually create task-specific data. This not only enhances the domain expertise of the dataset but also reduces the risk of data leakage.

\paragraph{Future Direction for Better Legal AI.}There are already many legal benchmarks available for individual countries—particularly for China, the United States, and Korea—but there is a lack of benchmarks that cover multiple languages or judicial domains, which are essential for evaluating multilingual LLMs. However, this is not merely about introducing additional languages into the legal domain; rather, it encourages researchers to consider real-world legal demands by developing more relevant and comprehensive task taxonomies—such as capturing the differences between Chinese and U.S. legal systems, or between civil law and common law traditions. Moreover, there is a lack of benchmarks addressing multimodal scenarios in the legal domain. Researchers should explore the multimodal demands of legal practice and leverage multimodal LLMs to meet these needs.

\subsubsection{Intellectual Property}

Intellectual property (IP) is an emerging field that is attracting increasing attention from natural language processing researchers. This is due to the diverse nature of intellectual property—including patents, copyrights, trademarks, and more—and its inherent legal significance in protecting the value and originality of creators’ work. In essence, intellectual property possesses a dual attribution: legal and technical. Among the various IP mechanisms, patents have been the most extensively researched. Before the emergence of LLMs, researchers primarily focused on three key areas: patent retrieval, patent classification, and patent generation. With the advancement of LLMs, processing the complex legal language of patents has become easier, enabling researchers to extend these tasks by leveraging LLM capabilities. 

PatentEval~\cite{PatentEval} focuses on patent generation tasks—particularly abstract generation and next-claim prediction—by evaluating models’ patent drafting capabilities and introducing a comprehensive error taxonomy. D2P~\cite{D2P} was introduced to generate complete, long-form patent documents from user-provided drafts, simulating real-world patent application scenarios. It also proposes a multi-agent framework, AutoPatent, to handle this complex generation task. Both of these benchmarks focus on evaluating the generation capabilities of LLMs in the intellectual property domain, where models must not only handle complex technical terminology and sentence structures, but also align with the legal language style and content specificity characteristic of IP documents. Both benchmarks aim to facilitate patent drafting and reduce the reliance on manual effort in these scenarios.

Beyond patent drafting and generation, there are also benchmarks designed to evaluate models’ capabilities in processing IP-related legal knowledge and technical content. MoZIP~\cite{MoZIP} is a multilingual benchmark covering ten different languages, comprising three tasks: IPQuiz, IPQA, and PatentMatch. The IPQuiz and IPQA tasks primarily focus on the legal aspects of intellectual property, aiming to evaluate models’ knowledge and their ability to apply it in practical IP scenarios. The PatentMatch task places greater emphasis on the technical aspects of patents, requiring the model to select the most similar patent from four carefully constructed candidates. A later literature, IPEval~\cite{IPEval}, focuses solely on the legal aspects of patents. It uses patent bar exam data in both Chinese and English, presented in a multiple-choice question answering format. 

The most recent study, IPBench~\cite{IPBench}, introduces the most comprehensive taxonomy in this domain with four hierarchical levels: \textbf{Information Processing}, \textbf{Logical Reasoning}, \textbf{Discriminant Evaluation}, and \textbf{Creative Generation}. This taxonomy is based on the American educational evaluator Norman L. Webb’s Depth of Knowledge Theory (DOK)~\cite{DOK}, which categorizes students’ cognitive levels into four tiers: Recall and Reproduction, Skill and Concept Application, Strategic Thinking, and Extended Thinking. From this cognitive perspective, it comprises 20 fine-grained tasks across the four levels, covering eight different IP mechanisms. These tasks—including multiple-choice question answering, patent classification, and generation—span a wide range of cognitive and practical demands, from IP-related legal memorization and interpretation to process guidance reflecting real-world practice. They also cover tasks such as infringement behavior determination and compensation calculation, requiring models not only to possess concrete legal knowledge, but also to demonstrate reasoning and mathematical computation abilities.

\paragraph{Patent Data Source.}Similar to legal benchmarks, the China National Intellectual Property Administration (CNIPA), the United States Patent and Trademark Office (USPTO), and the European Patent Office (EPO)—particularly the USPTO and EPO—provide patent access APIs. Additionally, the Google Patents Dataset offers relevant corpora. Benchmarks such as PatentEval~\cite{PatentEval}, D2P~\cite{D2P}, and MoZIP~\cite{MoZIP} utilize patent data from these sources. IPEval~\cite{IPEval} uses questions from national standard patent bar exams, while MoZIP’s IPQuiz and IPQA tasks collect FAQs from the official websites of IP organizations and agencies worldwide. In contrast, IPBench~\cite{IPBench} leverages extensive expert annotation to construct datasets that align with real-world application needs. All of these benchmarks are limited to the text modality, using only the textual portions of patents. However, patents also contain images, and domains such as trademarks and logos involve even richer visual information. Future research should focus on integrating and categorizing multimodal IP tasks to enable more comprehensive and intelligent IP services.

\subsubsection{Education}
Existing benchmarks such as GPQA~\cite{GPQA} and MMLU~\cite{MMLU} have been proposed to evaluate the knowledge levels of LLMs (e.g., at the graduate level). Although possessing relevant knowledge is a prerequisite for LLMs to support educational outcomes, these benchmarks differ from education-oriented ones, particularly in their alignment with real-world educational scenarios. Especially for kindergarten through twelfth grade (K–12) education in the Chinese context, E-Eval~\cite{E-EVAL} focuses on real-world classroom scenarios spanning primary, middle, and high school levels. It categorizes tasks into two main types: \textbf{arts} (including Chinese, English, and History) and \textbf{sciences} (such as Mathematics, Physics, and Chemistry). However, it is not sufficient for LLMs to merely know, understand, and apply knowledge. The education domain is complex and involves multiple stages, such as student preview, in-class teaching, and student ability assessment, among other concrete educational scenarios. These scenarios require dedicated design and exploration of effective methods to integrate LLMs into each stage of the educational process. 

EduBench~\cite{EduBench} takes into account the practical nature of the education domain, covering nine core educational scenarios and over 4,000 diverse synthetic education tasks. These tasks can be categorized into two types based on different teaching targets: \textbf{Student-Oriented Scenarios}, which include Problem Solving, Error Correction, Idea Provision, Personalized Learning Support, and Emotional Support; and \textbf{Teacher-Oriented Scenarios}, which include Question Generation, Automatic Grading, Teaching Material Generation, and Personalized Content Creation. These tasks are highly practical, but unlike E-Eval~\cite{E-EVAL}, which adopts a multiple-choice question answering format, they pose greater challenges in evaluating LLM performance within each specific scenario. For each scenario, EduBench~\cite{EduBench} evaluates whether LLMs can fulfill the teaching objectives and scenario-specific expectations across three dimensions: Scenario Adaptation, Factual \& Reasoning Accuracy, and Pedagogical Application, using DeepSeek-V3 as the evaluator.

Although EduBench~\cite{EduBench} provides a richer simulation of educational scenarios compared to E-Eval~\cite{E-EVAL}, making it more suitable for this domain, it offers only a coarse-grained framework. The education domain remains underexplored, and researchers need to focus on more fine-grained educational scenarios and investigate real-world teaching practices. This includes incorporating multimodal information and exploring how to better leverage LLMs to assist teachers in class preparation, as well as helping students learn effectively and correct their knowledge, ultimately benefiting their exam performance and personal knowledge framework construction.

\subsubsection{Psychology}
Recent literature has increasingly focused on human health in the context of using large language models (LLMs); however, most studies emphasize physical or biological aspects of health, with comparatively less attention given to mental health. Given the human-like conversational style of interactions with LLMs, it is possible for them to provide support for mental health, such as offering psychological insights and advice. CPsyExam~\cite{CPsyExam} is designed to benchmark the capabilities of LLMs in understanding psychological concepts through the Chinese Standard Examination.Its taxonomy includes two task types—\textbf{psychological knowledge} and \textbf{case analysis skills}—and incorporates three question formats: single-choice, multiple-choice, and question-answering. However, similar to the education domain, in this domain closely related to human interaction, it is important to explore concrete, context-specific practices. Relying solely on knowledge-based evaluation formats is insufficient to deeply assess LLMs’ capabilities in the psychological domain. Motivated by this, Zhang et al.~\cite{CPsyCoun} propose CPSYCOUN, a benchmark aimed at evaluating multi-turn dialogues between humans and LLMs to explore whether LLMs can be effectively applied in Chinese psychological counseling scenarios. The benchmark covers nine topics and seven classic schools of psychological counseling, including Psychoanalytic Therapy and Cognitive Behavioral Therapy. Although CPsyCoun~\cite{CPsyCoun} recognizes the practical need to use LLMs for serving people and evaluating them in psychological settings, it still lacks a more fine-grained taxonomy of psychological scenes and tasks. Moreover, both CPsyExam and CPsyCoun focus primarily on the Chinese psychological context, leaving a gap in multilingual benchmarks for this human-centric domain. Psycollm~\cite{Psycollm} also focuses on Chinese, constructing a benchmark based on authoritative psychological counseling examinations in China, using both single-turn and multi-turn question answering formats, including assessments
of \textbf{professional ethics}, \textbf{theoretical proficiency}, and \textbf{case analysis}. These types of questions are highly related to the scenarios. 

Beyond the psychological services, many literatures explore whether LLMs possess attributes similar to those of humans and examines the stability of these attributes. Psychometrics Benchmark~\cite{li2024quantifying} introduces a framework leveraged to measure LLMs’ psychological attributions, covering six dimensions: personality, values, emotion, theory of mind, motivation, and intelligence. PsychoBench~\cite{PsychBench} also focuses on LLMs’ personality, temperament, and emotion. Using thirteen scales commonly applied in clinical psychology, PsychoBench further classifies these scales into four categories: \textbf{personality traits}, \textbf{interpersonal relationships}, \textbf{motivational tests}, and \textbf{emotional abilities}. Both benchmarks use scenario-based scales to quantify models’ psychological characteristics.

\subsubsection{Finance}
Financial technology has advanced alongside the development of language models. Before the surge of LLMs, many finance-specific models were built on BERT as their backbone, such as FinBERT and FLANG. BBT-CFLEB~\cite{BBT-CFLEB} was introduced to benchmark model performance—particularly that of BERT-based models—using data primarily composed of financial news and social media content. Although LLM performance on BBT-CFLEB has not been reported, it represents an important component of financial news analysis, which involves many domain-specific terms. PIXIU~\cite{FLARE} is a benchmark specifically designed for financial LLMs. It includes not only NLP-related tasks but also financial prediction tasks such as stock movement prediction, which require deeper domain-specific knowledge. As a real-world and practice-oriented domain, finance also contains many scenarios that remain to be explored. FinEval~\cite{Fineval} categorizes financial knowledge and practical abilities into four key types: \textbf{Financial Academic Knowledge}, \textbf{Financial Industry Knowledge}, \textbf{Financial Security Knowledge}, and \textbf{Financial Agent}. These categories cover areas such as financial and economic knowledge, the use of financial tools, and financial reasoning. All of these financial benchmarks are monolingual, highlighting the need for researchers to explore financial environments across different countries and languages.

\subsubsection{Summary and Future Directions}
\paragraph{Fine-grained Task Taxonomy.}All these human and social science domains are highly grounded in real-world practice and hold significant practical value, especially for individuals. This characteristic makes the definition of concrete, scenario-based tasks particularly important. Therefore, before evaluating the capabilities of LLMs in these domains—especially in finance, psychology, and education—it is essential to establish a fine-grained task taxonomy.

\paragraph{Developing Robust Methods for Evaluation.}In domains with rich real-world applications, relying solely on multiple-choice questions is insufficient for evaluation. There is an urgent need to develop scenario-oriented evaluation methods to better assess the concrete performance of language models. This necessitates that researchers, grounded in a fine-grained taxonomy, design evaluation formats that are more detailed, concrete, and robust for these domains.

\subsection{Engineering \& Technology}

The engineering and technology domain represents a crucible for Large Language Models, testing their capabilities on tasks that demand not only linguistic fluency but also logical rigor, functional correctness, and deep, specialized knowledge. Unlike general-purpose tasks, engineering applications often have a single correct answer or a narrow range of acceptable solutions, governed by physical laws, mathematical principles, or strict syntax. Success in this area requires models to act as functional tools rather than just conversational partners. Consequently, this field has fostered some of the most sophisticated and mature evaluation frameworks. This section surveys the landscape of engineering benchmarks, tracing their evolution from foundational code generation to complex, multi-domain problem-solving across software, electrical, mechanical, and other engineering disciplines. A summary of representative benchmarks is provided in Table \ref{tab:engineering_benchmarks}.

\begin{table}[h!]
\centering
\renewcommand{\arraystretch}{1}
\resizebox{\textwidth}{!}{
\begin{tabular}{llllllllll}
\hline
\textbf{Benchmark} & \textbf{Focus} & \textbf{Language} & \textbf{Source} & \textbf{Data Type} & \textbf{Eval.} & \textbf{Indicators} & \textbf{Amount} & \textbf{Method} & \textbf{Citations} \\
\hline
\multicolumn{10}{l}{\textbf{\textit{Software Engineering \& Information Technology}}} \\
HumanEval \cite{HumanEval} & Code Generation & Monolingual & Manual Design & Generation & AE & Pass@k & 164 & No & 4970 \\
MBPP \cite{MBPP} & Python Code Gen & Monolingual & Manual Design & Generation & AE & Pass@k & 974 & No & 1935 \\
SWE-bench \cite{SWE-bench} & \begin{tabular}[t]{@{}l@{}}GitHub Issue \\ Repair\end{tabular} & Monolingual & Open Datasets & Generation & AE & Pass Rate & 2294 & No & 680 \\
LiveCodeBench \cite{LiveCodeBench} & \begin{tabular}[t]{@{}l@{}}Live Contest \\ Programming\end{tabular} & Monolingual & Web & Generation & AE & Pass@1 & 511 & Yes & 401 \\
ClassEval \cite{ClassEval} & Class-level Gen. & Monolingual & Manual Design & Generation & AE & \begin{tabular}[t]{@{}l@{}}Pass@k, DET\end{tabular} & 100 & Yes & 154 \\
xCodeEval \cite{xCodeEval} & Multilingual Tasks & Multilingual & Web & Hybrid & AE & \begin{tabular}[t]{@{}l@{}}Pass@k, \\ macro-F1\end{tabular} & 7514 & Yes & 53 \\
CodeReview \cite{CodeReview} & Code Review & Multilingual & Web & Generation & ME & \begin{tabular}[t]{@{}l@{}}Acc, BLEU, \\ EM\end{tabular} & 534k & No & 202 \\
Spider \cite{Spider} & Cross-Domain SQL & Monolingual & Hybrid & Generation & AE & EM, EX & 10k & No & 1471 \\
BIRD \cite{BIRD} & Large-scale SQL & Monolingual & Hybrid & Generation & AE & Execution Acc & 12.7k & No & 479 \\
CoSQL \cite{CoSQL} & Conversational SQL & Monolingual & Hybrid & Generation & AE & Acc, BLEU & 3k & No & 110 \\
IaC-Eval \cite{IaC-Eval} & IaC Generation & Monolingual & Manual Design & Generation & AE & Pass@k & 458 & Yes & 10 \\
OpsEval \cite{OpsEval} & AIOps & Bilingual & Hybrid & Hybrid & ME & FAE-Score & 9070 & Yes & 7 \\
FrontendBench \cite{FrontendBench} & Frontend Dev. & Monolingual & Hybrid & Generation & AE & \begin{tabular}[t]{@{}l@{}}PassRate, \\ Consistency\end{tabular} & 148 pairs & Yes & 0 \\
\hline
\multicolumn{10}{l}{\textbf{\textit{Electrical \& Electronic Engineering}}} \\
VerilogEval \cite{VerilogEval} & Verilog Generation & Monolingual & Hybrid & Generation & AE & Pass@k & 156 & Yes & 218 \\
RTLLM \cite{RTLLM} & RTL Generation & Monolingual & Manual Design & Generation & AE & \begin{tabular}[t]{@{}l@{}}Syntax/Func, \\ Quality\end{tabular} & 30 & Yes & 182 \\
CIRCUIT \cite{CIRCUIT} & Analog Circuit & Monolingual & Manual Design & Generation & ME & \begin{tabular}[t]{@{}l@{}}Global Acc, \\ Template Acc\end{tabular} & 510 pairs & Yes & 0 \\
ElecBench \cite{ElecBench} & Power Dispatch & Monolingual & Hybrid & Hybrid & ME & \begin{tabular}[t]{@{}l@{}}Authenticity, \\ Logicality, Stability, \\ Security, etc.\end{tabular} & N/A & Yes & 0 \\
FIXME \cite{FIXME} & HW Verification & Monolingual & Hybrid & Hybrid & AE & \begin{tabular}[t]{@{}l@{}}Pass rate, \\ Correctness\end{tabular} & 180 tasks & Yes & 0 \\
\hline
\multicolumn{10}{l}{\textbf{\textit{Mechanical, Manufacturing, Aerospace \& Transportation Engineering}}} \\
CADBench \cite{CADBench} & CAD Script Gen. & Monolingual & Hybrid & Generation & ME & \begin{tabular}[t]{@{}l@{}}Avg scores, \\ Syntax error\end{tabular} & 700 & Yes & 0 \\
LLM4Mat-bench \cite{LLM4Mat-bench} & Material Property & Monolingual & Web & Hybrid & AE & \begin{tabular}[t]{@{}l@{}}MAD:MAE ratio, \\ AUC\end{tabular} & 1.98M & Yes & 14 \\
AeroMfg-QA \cite{AeroManufacturing-QA} & Aerospace Mfg. & Monolingual & Manual Design & MCQA & AE & \begin{tabular}[t]{@{}l@{}}Custom score, \\ Acc\end{tabular} & 2480 & Yes & 0 \\
RepoSpace \cite{RepoSpace} & Aerospace Repo. & Monolingual & Private & Generation & AE & \begin{tabular}[t]{@{}l@{}}Rouge-L, \\ CodeBLEU\end{tabular} & 825 & Yes & 0 \\
Aviation-Benchmark \cite{Aviation-Benchmark} & Aviation Domain & Monolingual & Open Datasets & Hybrid & AE & Accuracy & 150k & Yes & 0 \\
\hline
\end{tabular}
}
\caption{Summary of representative benchmarks in Engineering \& Technology. AE: Automated Evaluation; ME: Mixed Evaluation; Gen.: Generation. The 'Method' column indicates if the paper proposed a new methodology (Yes/No).}
\label{tab:engineering_benchmarks}
\end{table}

\subsubsection{Software Engineering and Information Technology}
As the discipline most intertwined with the development of AI, software engineering has the most extensive and mature collection of benchmarks. These evaluations span the entire software development lifecycle, from initial ideation to long-term maintenance.

\paragraph{Software Development and Maintenance} The journey of evaluation began with foundational \textbf{code generation} tasks. Benchmarks like HumanEval \cite{HumanEval} and MBPP \cite{MBPP} established the now-standard paradigm of assessing function-level code synthesis from natural language prompts, using functional correctness (pass@k) via unit tests as the primary metric. This initial focus quickly expanded to address greater complexity and realism. For instance, APPS \cite{APPS} and USACO \cite{USACO} introduced problems from programming competitions, demanding more advanced algorithmic reasoning. To combat the pervasive issue of benchmark contamination, LiveCodeBench \cite{LiveCodeBench} and its expert-level successor LiveCodeBenchPro \cite{LiveCodeBenchPro} pioneered the use of problems from live, ongoing contests, ensuring that models are evaluated on truly unseen data.

The scope of generation tasks has also broadened from simple, self-contained functions to more complex software artifacts. ClassEval \cite{ClassEval} was the first to specifically target class-level generation, a crucial step for evaluating object-oriented programming skills. The evaluation of domain-specific code generation has also become a major trend, with benchmarks like DS-1000 \cite{DS-1000} for data science libraries, BioCoder \cite{BioCoder} for bioinformatics, and the recent MMCode \cite{MMCode}, which challenges models with multimodal problems containing visual information like diagrams.

Beyond initial creation, a model's ability to work with existing code is critical. Comprehensive frameworks for \textbf{code understanding and completion}, such as CodeXGLUE \cite{CodeXGLUE} and the multilingual xCodeEval \cite{xCodeEval}, offer a suite of tasks including code summarization, translation, and retrieval. Other benchmarks target more specific understanding tasks, such as code question-answering (CodeQA \cite{CodeQA}), code search within large corpora (Cosqa \cite{Cosqa}), and repository-level code completion that mimics a developer's IDE experience (Repobench \cite{Repobench}).

\textbf{Code maintenance}, a significant portion of real-world software engineering, is another vital evaluation area. Automated program repair is a key focus, with benchmarks like RepairBench \cite{RepairBench} and the highly influential SWE-bench \cite{SWE-bench}. The latter is particularly notable for sourcing its tasks directly from real GitHub issues and pull requests in popular open-source projects, providing an unparalleled level of realism. Complementary benchmarks like Debugbench \cite{Debugbench} and Condefects \cite{Condefects} focus on the related skills of debugging and defect localization. Furthermore, precise code editing is evaluated by CanItEdit \cite{CanItEdit} and CodeEditorBench \cite{CodeEditorBench}, while code efficiency—a crucial non-functional requirement—is measured by benchmarks such as COFFE \cite{COFFE} and EffiBench \cite{EffiBench}.

\paragraph{Database Systems and DevOps} In the realm of database systems, Text-to-SQL translation remains the predominant evaluation task, as it is key to democratizing data access. The Spider \cite{Spider} benchmark is the established standard for complex, cross-domain queries. Its successors, such as Spider 2.0 \cite{Spider2} and BIRD \cite{BIRD}, have increased the difficulty by incorporating more realistic enterprise workflows and value-based queries. The evaluation has also evolved to include conversational context, where models must understand multi-turn dialogues (CoSQL \cite{CoSQL}, SParC \cite{SParC}), handle robustness against perturbations (Dr. Spider \cite{DrSpider}), and support multilingual queries (DuSQL \cite{DuSQL}).

For System Administration and DevOps, benchmarks are emerging to assess the automation of operational tasks. This includes translating natural language to shell commands (NL2Bash \cite{NL2Bash}), generating Infrastructure-as-Code (IaC) for cloud services (IaC-Eval \cite{IaC-Eval}), and solving broader AIOps problems (OpsEval \cite{OpsEval}, OWL \cite{OWL-Bench}). In Human-Computer Interaction, FrontendBench \cite{FrontendBench} specifically evaluates the generation of code for interactive web user interfaces.

\subsubsection{Specialized Engineering Disciplines}
Beyond software, evaluation frameworks are being developed for hardware and physical engineering domains, which introduce challenges related to physical laws, safety constraints, and highly specialized languages.

In \textbf{Electrical and Electronic Engineering}, benchmarks for chip design automation are a primary focus. VerilogEval \cite{VerilogEval} and RTLLM \cite{RTLLM} assess the ability to generate Hardware Description Languages (HDL) like Verilog, a critical skill for designing digital integrated circuits. The evaluation goes beyond mere syntax to include functional correctness through simulation. The focus also extends to the efficiency of the generated hardware, with ResBench \cite{ResBench} measuring FPGA resource utilization, and to the crucial task of design verification, with FIXME \cite{FIXME} providing an end-to-end framework. Other specialized areas include analog circuit design, assessed by CIRCUIT \cite{CIRCUIT}, and the niche field of photonic circuits, covered by PICBench \cite{PICBench}. In the domain of power systems, ElecBench \cite{ElecBench} evaluates LLM performance on complex power dispatch and fault diagnosis tasks.

In \textbf{Mechanical and Manufacturing Engineering}, benchmarks like CADBench \cite{CADBench} assess the generation of scripts for Computer-Aided Design (CAD) software, a key task in automating mechanical design. Materials science is another active area, where benchmarks like LLM4Mat-bench \cite{LLM4Mat-bench} and MSQA \cite{MSQA} test the ability of LLMs to accelerate materials discovery by predicting chemical properties and demonstrating graduate-level reasoning.

For \textbf{Aerospace and Transportation Engineering}, a safety-critical domain, specialized benchmarks have been developed to evaluate knowledge and code generation. These include AeroManufacturing-QA \cite{AeroManufacturing-QA} for assessing expertise in aerospace manufacturing processes, RepoSpace \cite{RepoSpace} for evaluating repository-level code generation for satellite systems, and the broad Aviation-Benchmark \cite{Aviation-Benchmark} which covers over ten specific aviation tasks.

\subsubsection{Summary and Future Directions}
The engineering domain has driven the development of some of the most rigorous, functionally-grounded, and execution-based evaluations for LLMs. The clear trend is a progression from assessing isolated, function-level skills towards evaluating performance on complex, system-level problems that mirror complete engineering workflows. Despite this progress, a significant gap persists between high performance on benchmarks and reliable deployment in real-world, mission-critical engineering applications.

Future research in this area must prioritize several key directions. First, there is a pressing need for benchmarks that can evaluate \textbf{holistic engineering workflows}, integrating tasks from requirements analysis and high-level design through to implementation, verification, and long-term maintenance. Second, establishing robust and standardized evaluation protocols for \textbf{safety, reliability, and security} is paramount, especially for domains where failure can have catastrophic consequences. Third, the community must continue to develop \textbf{dynamic and contamination-resistant benchmarks} to ensure that evaluations remain a true test of generalization for ever-more-powerful models. Finally, as LLMs become integrated into engineering teams, it is crucial to develop frameworks that assess \textbf{human-AI collaboration}, measuring metrics like efficiency gains, error reduction, and trust calibration, rather than evaluating the AI in isolation. Addressing these challenges will be essential for unlocking the full potential of LLMs as transformative tools in engineering practice.

\section{Target-specific benchmarks}

\subsection{Risk \& Reliability}
\begin{table}[h!]
\centering
\renewcommand{\arraystretch}{1}
\resizebox{\textwidth}{!}{
\begin{tabular}{llllllllll}
\hline
Benchmark & Focus & Language & Source & Data Type & Eval. & Indicators & Amount & Method & Citations \\ 
\hline
\multicolumn{10}{l}{\textbf{\textit{Safety}}} \\
StereoSet\cite{StereoSet} & Safety & Monolingual & Manual Design & MCQA & Automated & \begin{tabular}[t]{@{}l@{}}LM Score\\SS Score\\ICAT Score\end{tabular} & 2.12k & No & 1238 \\
CrowS-Pairs\cite{CrowS-Pairs} & Safety & Bilingual  & Manual Design & MCQA & Automated & \begin{tabular}[t]{@{}l@{}}Bias Score\end{tabular} & 1,508 & No & 821 \\
HateCheck\cite{HateCheck} & Safety & Multilingual & Manual Design & Classification & Automated & \begin{tabular}[t]{@{}l@{}}Accuracy\end{tabular} & 3,728 & Yes & 313 \\
ToxiGen\cite{ToxiGen} & Safety & Monolingual & Model Generation & Classification & Automated & \begin{tabular}[t]{@{}l@{}}AUC\end{tabular} & 274,186 & No & 547 \\
Do-Not-Answer\cite{Do-Not-Answer} & Safety & Bilingual & Manual Design & Classification & Hybrid & Accuracy, Precision, Recall, F1 & 939 & No & 133 \\
SG-Bench\cite{SG-Bench} & Safety & Monolingual & Hybrid & Hybrid & Hybrid & Failure Rate & 1442 & Yes & 10 \\
JailbreakBench\cite{JailbreakBench} & Safety & Monolingual & Hybrid & Generation & Hybrid & \begin{tabular}[t]{@{}l@{}}Jailbreak Success Rate\\Defense Effectiveness\end{tabular} & 200 & No & 244 \\
AnswerCarefully\cite{AnswerCarefully} & Safety & Bilingual  & Manual Design & Generation & Automated & \begin{tabular}[t]{@{}l@{}}Violation Rate\\Acceptable Response Rate\end{tabular} & 1800 & No & 2 \\
SorryBench\cite{SorryBench} & Safety & Multilingual  & Hybrid & Generation & Automated & fulfillment Rate & 9240 & No & 120 \\
MaliciousInstruct\cite{MaliciousInstruct} & Safety & Monolingual  & Hybrid & Generation & Automated & \begin{tabular}[t]{@{}l@{}}Attack Success Rate\\Harmfulness Percentage\end{tabular} & 100 & No & 362 \\
HarmBench\cite{HarmBench} & Safety & Monolingual  & Manual Design & Generation & Automated & Attack Success Rate & 500 & Yes & 434 \\
HEx-PHI\cite{HEx-PHI} & Safety & Monolingual  & Manual Design & Generation & Automated & \begin{tabular}[t]{@{}l@{}}Harmfulness Score\\Harmfulness Rate\end{tabular} & 330 & No & 722 \\
SimpleSafetyTests\cite{Simplesafetytests} & Safety & Monolingual  & Hybrid & Classification & Automated & Accuracy & 3000 & No & 48 \\
ToxicChat\cite{ToxicChat} & Safety & Monolingual  & Hybrid & Generation & Human & \begin{tabular}[t]{@{}l@{}}Precision, Recall, F1\\jalibreaking recall\end{tabular} & 10166 & No & 139 \\
In-The-Wild Jailbreak Prompts\cite{In-The-Wild} & Safety & Monolingual  & Hybrid & Generation & Automated & Attack Success Rate & 107250 & No & 721 \\
\hline
\multicolumn{10}{l}{\textbf{\textit{Hallucination}}} \\
TruthfulQA\cite{TruthfulQA} & Hallucination & Multilingual & Manual Design & Generation & Human& \begin{tabular}[t]{@{}l@{}}Truthfulness Score\end{tabular} & 817 & No & 2098 \\
FActScore\cite{FActScore} & Hallucination & Multilingual & Open Datasets & Generation & Hybrid & \begin{tabular}[t]{@{}l@{}}FActScore\end{tabular} & 683 & Yes & 707 \\
RealtimeQA\cite{RealtimeQA} & Hallucination & Monolingual & Web & Generation & Hybrid & \begin{tabular}[t]{@{}l@{}}Real-time Accuracy\end{tabular} & 30 & Yes & 168 \\
FaithBench\cite{FaithBench} & Hallucination & Monolingual & Hybrid & Generation & Human& \begin{tabular}[t]{@{}l@{}}Faithfulness Score\end{tabular} & 800 & Yes & 8 \\
DiaHalu\cite{DiaHalu} & Hallucination & Monolingual & Model Generation & Generation & Human& \begin{tabular}[t]{@{}l@{}}Hallucination Rate\end{tabular} & 1103 & Yes & 15 \\
FactCheck-Bench\cite{FactCheckBench} & Hallucination & Monolingual & Model Generation & Hybrid & Hybrid & \begin{tabular}[t]{@{}l@{}}Fact-check Accuracy\end{tabular} & 94 & Yes & 39 \\
FELM\cite{FELM} & Hallucination & Monolingual & Hybrid & Generation & Hybrid & \begin{tabular}[t]{@{}l@{}}Factual Consistency Score\end{tabular} & 847 & Yes & 106\\
FACTOR\cite{FACTOR} & Hallucination & Monolingual & Open Datasets & MCQA & Hybrid & \begin{tabular}[t]{@{}l@{}}Factual Accuracy\end{tabular} & 4266 & Yes & 106 \\
FreshQA\cite{FreshQA} & Hallucination & Monolingual & Manual Design & Generation & Hybrid & \begin{tabular}[t]{@{}l@{}}Freshness Score\end{tabular} & 600 & Yes & 246 \\
MedHallu\cite{MedHallu} & Hallucination & Monolingual & Hybrid & Classification & Human& \begin{tabular}[t]{@{}l@{}}Medical Accuracy\end{tabular} & 10000 & Yes & 5 \\
HaluEval\cite{HaluEval} & Hallucination & Monolingual & Hybrid & Generation & Hybrid & \begin{tabular}[t]{@{}l@{}}Hallucination Rate\end{tabular} & 35000 & No & 533 \\
HaluEval2.0\cite{HaluEval2.0} & Hallucination & Monolingual & Hybrid & Generation & Hybrid & \begin{tabular}[t]{@{}l@{}}Hallucination Rate\end{tabular} & 8770 & No & 138 \\
FaithDial\cite{FaithDial} & Hallucination & Monolingual & Hybrid & Generation & Human& \begin{tabular}[t]{@{}l@{}}Faithfulness Score\end{tabular} & 5649 & No & 100 \\
\hline
\multicolumn{10}{l}{\textbf{\textit{Robustness}}} \\
AdvGLUE\cite{AdvGLUE} & Robustness & Monolingual & Hybrid & Classification & Automated & \begin{tabular}[t]{@{}l@{}}Robustness Score\end{tabular} & 5716 & No & 284 \\
BOSS\cite{BOSS} & Robustness & Monolingual & Open Datasets & Classification & Automated & \begin{tabular}[t]{@{}l@{}}Robustness Score\end{tabular} & 900 & No & 8 \\
IFEval\cite{IFEval} & Robustness & Monolingual & Manual Design & Generation & Automated & \begin{tabular}[t]{@{}l@{}}Inference Robustness\end{tabular} & 541 & Yes & 438 \\
CIF-Bench\cite{CIF-Bench} & Robustness & Monolingual & Hybrid & Generation & Automated & \begin{tabular}[t]{@{}l@{}}Consistency Score\end{tabular} & 45000 & No & 16 \\
PromptRobust\cite{PromptRobust} & Robustness & Monolingual & Hybrid & Hybrid & Automated & \begin{tabular}[t]{@{}l@{}}Robustness Score\end{tabular} & 4788 & Yes & 52 \\
RoTBench\cite{RoTBench} & Robustness & Monolingual & Manual Design & Generation & Automated & \begin{tabular}[t]{@{}l@{}}Robustness Score\end{tabular} & 4077 & Yes & 18 \\
\hline
\multicolumn{10}{l}{\textbf{\textit{Data Leak}}} \\
WikiMIA \cite{WikiMIA} & Data Leak & Monolingual & Open Datasets & Classification & Automated & \begin{tabular}[t]{@{}l@{}}Min-k\% Prob\end{tabular} & 250 & Yes & 368 \\
KoLA\cite{KoLA} & Data Leak & Bilingual & Hybrid & Generation & Automated & \begin{tabular}[t]{@{}l@{}}Accuracy\end{tabular} & 500 & No & 144 \\
C$^2$LEVA\cite{C2LEVA} & Data Leak & Bilingual & Hybrid & Hybrid & Automated & \begin{tabular}[t]{@{}l@{}}Mean Win Rate\end{tabular} & 16115 & Yes & 8 \\
\hline
\end{tabular}
}
\caption{Comprehensive Summary of Risk \& Reliability Benchmarks. The 'Method' column indicates if the paper proposed a new methodology (Yes/No).}
\label{tab:risk_reliability_benchmarks}
\end{table}

The rapid advancement of Large Language Models (LLMs)\cite{gpt4, qwen3, kimik2} has unlocked unprecedented potential across a wide spectrum of applications. However, as these models transition from research prototypes to real-world deployment, particularly in high-stakes scenarios such as medical consultation, legal reasoning, financial advising, or customer support, their immense capabilities are shadowed by equally significant risks. Issues like hallucinations, biased outputs, adversarial susceptibility, and privacy violations are no longer theoretical, they have tangible consequences for users, organizations, and society at large.

Consequently, Risk \& Reliability assessment has evolved into a central pillar of modern LLM benchmarking frameworks, rather than a peripheral addition. Its core motivations are:

\textbf{1. Identification and Quantification}: To systematically probe LLMs for various negative impact patterns (e.g., generating harmful content\cite{safetyRLHF}, hallucinating facts\cite{HaluSurvey}, leaking private data\cite{ProPILE}) and quantify the frequency and severity of these risks. This necessitates testing under diverse, challenging inputs, including extremes, adversarial prompts, and edge cases (e.g., jailbreak attempts\cite{jailbreaksurvey}, biased prompts\cite{ToxiGen}, fact-intensive queries\cite{TruthfulQA}).

\textbf{2. Risk Mitigation}: To utilize the weaknesses revealed by benchmarks to drive technical improvements (e.g., more robust RLHF, factuality enhancement, privacy-preserving training) by developers, and inform more effective safeguards (e.g., content filters, usage policies) for deployers. The ultimate goal is to minimize the likelihood of models malfunctioning or causing harm.

\textbf{3. Alignment with Expectations}: To verify that model behavior adheres to predefined ethical principles, legal boundaries, and safety requirements (i.e., the alignment problem) during complex, real-world interactions, demonstrating robustness especially on sensitive topics.

\textbf{4. Building and Sustaining Trust}: To provide rigorous, reproducible evidence of risks to demonstrate to users, regulators, and society that a specific LLM is sufficiently reliable, safe, and trustworthy, thereby fostering healthy ecosystem growth and responsible widespread adoption.

In essence, the core question this research direction addresses is: \textbf{Beyond impressive capabilities, is this model safe, reliable, and trustworthy enough?} It aims to supply the empirical foundation for the liability guarantee of the model, serving as the essential security checkpoint for LLMs transitioning from research labs to the real world.

\subsubsection{Safety}

Following pre-training, large language models (LLMs) rely on safety alignment to balance helpfulness and harmlessness. However, ensuring harmlessness often necessitates imposing strict constraints on the model's output space, creating a fundamental conflict with their core strength of deep instruction-following capability, whose objective is to broadly understand and respond to user requests. The inherent flaws in pre-training data, namely the inevitable inclusion of harmful content—embed a latent vulnerability within this conflict: users can employ "jailbreak" techniques to subtly activate these remnants of harmful knowledge retained internally by the model, thereby compromising its safety barriers. Particularly noteworthy is that the explicit reasoning chain approach adopted starting with models like GPT o1, while enhancing interpretability, also unintentionally introduces an observable and steerable pathway for jailbreak attacks. This amplifies safety risks, creating new challenges for the reliable deployment of these models.

Early studies such as HateCheck\cite{HateCheck}, StereoSet\cite{StereoSet}, and CrowS-Pairs\cite{CrowS-Pairs} primarily relied on predefined harmful scenarios and static test cases to evaluate model safety capabilities. However, these approaches suffered from an over-reliance on manually constructed datasets, resulting in limited coverage. Addressing this limitation, ToxiGen\cite{ToxiGen} leveraged large language models to generate large-scale adversarial and implicit harmful content (up to 274K samples), significantly enhancing the scale and complexity of test sets. This advancement facilitated the development of models with improved generalization capabilities in real-world scenarios. Expanding evaluation dimensions further, Do-Not-Answer\cite{Do-Not-Answer} addressed the gap in safety assessment for Chinese-language contexts by establishing a standardized testing framework covering eight categories of sensitive topics, including healthcare and criminal activities. To confront emerging challenges in model safety defenses, JailbreakBench\cite{JailbreakBench} systematically integrated over a hundred adversarial prompt techniques (e.g., role-playing and logic-exploiting prompts) to diagnose model vulnerabilities. Building upon this foundation, SG-Bench\cite{SG-Bench} introduced a cross-task safety generalization evaluation framework designed to test robustness against unseen attack patterns, shifting the evaluation paradigm from static testing toward complex, dynamic interactions. AnswerCarefully\cite{AnswerCarefully} further broadened the scope by focusing on Japanese-language contexts, offering 1,800 carefully curated question–answer pairs aligned with cultural norms, thus serving both instruction tuning and safety validation.

More recently, additional benchmarks have expanded the landscape of safety evaluations. HarmBench\cite{HarmBench} introduced the first standardized framework for automated red-teaming and robustness refusal evaluation, covering 510 unique harmful behaviors across text and multimodal settings. It also proposed the R2D2 dynamic defense method and conducted the largest comparative study to date, evaluating 18 attack methods across 33 models. HEx-PHI\cite{HEx-PHI} highlighted security risks introduced by fine-tuning, even in non-malicious contexts, and constructed a dataset of 330 red-team prompts grounded in usage policies from Meta and OpenAI to assess vulnerabilities across prohibited use categories. SimpleSafetyTests\cite{Simplesafetytests} provided a lightweight suite of 100 English prompts spanning five high-risk domains (e.g., self-harm, scams, child abuse) and demonstrated the effectiveness of lightweight safety filters, with GPT-4-based content moderation achieving the highest accuracy.

Other datasets focus on real-world user interactions. ToxicChat\cite{ToxicChat} introduced 10,166 toxicity-labeled samples collected from real user–AI conversations with Vicuna, including explicit “jailbreaking recall” metrics to capture hidden adversarial attempts. In-the-wild Jailbreak Prompts\cite{In-The-Wild} systematically studied 1,405 jailbreak prompts gathered from Reddit, Discord, websites, and open datasets, showing their high attack success rates across major LLMs. Similarly, MaliciousInstruct\cite{MaliciousInstruct} exposed the fragility of alignment in open-source LLMs by demonstrating that simple variations in decoding strategies could raise attack success rates from 0\% to over 95\%, revealing serious weaknesses in current alignment pipelines. Finally, SORRY-Bench\cite{SorryBench} focused on safety refusal evaluation, providing a fine-grained taxonomy of 44 unsafe categories and 440 balanced prompts, along with 20 linguistic augmentations to test cross-linguistic and formatting robustness. Its human-in-the-loop design and efficient LLM-as-a-judge framework allow accurate, large-scale safety refusal benchmarking at lower computational cost.

Taken together, these benchmarks have significantly advanced the study of model safety by broadening the scope beyond purely static and English-only datasets toward more multilingual, adversarial, and fine-grained evaluations. Nonetheless, most current efforts remain grounded in static test suites, and truly dynamic evaluations—those that capture evolving attack strategies and interactive failure modes—are still underexplored. As such, developing systematic dynamic safety benchmarks remains an important direction for future work, and is essential for building safer and more reliable large language models.

\subsubsection{Hallucination}
Current large language models (LLMs) face the problem of hallucinations, which are primarily categorized into two types: \textbf{factual hallucinations}, where the model output contradicts verifiable facts, manifesting as factual inconsistency or fabrication, and \textbf{faithfulness hallucinations}, where outputs deviate from user instructions, input context, or lack internal logical consistency.

The causes of hallucinations span the entire model lifecycle. At the data level, misinformation, domain gaps, outdated knowledge, and deficiencies in rare knowledge recall and reasoning contribute significantly. At the training level, limitations such as unidirectional attention’s poor contextual capture, the mismatch between autoregressive training and inference, and alignment phase issues like capability or belief misalignment exacerbate the problem. During inference, randomness introduced by decoding strategies (e.g., high-temperature sampling) and architectural constraints (e.g., attention locality and the Softmax bottleneck) further distort factual fidelity.

To address these challenges, primary solutions include data cleaning, retrieval-augmented generation (RAG), and knowledge editing for knowledge enhancement. At the model level, improvements in architecture and alignment data preparation are crucial. During inference, decoding strategies designed to enhance factuality and logical consistency can significantly reduce hallucinations.

To systematically evaluate these hallucinations, a diverse set of benchmarks has been developed. These benchmarks vary in scope, language coverage, task format, and annotation methodology:

TruthfulQA\cite{TruthfulQA} identifies hallucinations where models mimic common human misconceptions across domains like science and history. FActScore\cite{FActScore} evaluates factual grounding in long-form generation by decomposing outputs into atomic facts (e.g., entities and events) and verifying them against external knowledge sources. REALTIMEQA\cite{RealtimeQA} focuses on hallucinations arising from stale knowledge, testing LLMs' adaptability to dynamic, real-time information (e.g., sports, finance).

For faithfulness distortions, FaithBench\cite{FaithBench} detects whether summarizations introduce information absent in source texts. DiaHalu\cite{DiaHalu} targets contextual contradictions in multi-turn dialogues, identifying issues like broken causality or entity inconsistency. FaithDial\cite{FaithDial} further expands this by evaluating dialogue systems on their fidelity to input conversational context.

Several benchmarks focus on domain-specific or adversarial hallucinations. MedHallu\cite{MedHallu} addresses hallucinations in medical generation tasks, ensuring alignment with trusted clinical knowledge. FreshQA\cite{FreshQA} evaluates model freshness by probing with up-to-date world knowledge. FACTOR\cite{FACTOR} introduces adversarial conditions (e.g., conflicting prompts) to test models' ability to resist and correct factual errors in real time.

Tools such as FELM\cite{FELM} extend beyond detection by requiring traceable correction paths and explanation-based justifications. FactCheck-Bench\cite{FactCheckBench} incorporates both model-generated and manually curated samples to measure fact-checking accuracy. Large-scale datasets like HaluEval\cite{HaluEval} and its improved variant HaluEval2.0\cite{HaluEval2.0} provide broad coverage across summarization, dialogue, and question answering, enabling benchmarking of hallucination rate at scale.

Despite the diversity of tools, the field still faces three persistent challenges: (1) the lack of a unified evaluation framework leads to fragmented coverage; (2) long-document coherence hallucinations remain difficult to detect; and (3) definitional ambiguity persists when differentiating between subjective judgments and verifiable facts.

\subsubsection{Robustness}

The rapid advancement of Large Language Models (LLMs) has significantly enhanced the capabilities of natural language processing systems. However, these models often exhibit vulnerabilities when exposed to adversarial inputs, distributional shifts, or subtle prompt variations. Such fragility can lead to erroneous or biased outputs, posing risks in critical applications. Consequently, evaluating and enhancing the robustness of LLMs has become a pivotal research focus.

Robustness in LLMs encompasses several dimensions. Adversarial robustness assesses model resilience against intentionally crafted inputs designed to mislead or deceive. Instruction-following robustness evaluates the consistency and accuracy of models in adhering to varied or complex instructions. Prompt robustness measures sensitivity to minor changes in prompt phrasing or structure. Tool-use robustness determines stability in scenarios requiring external tool integration or multi-step reasoning. These categories guide the development of benchmarks aimed at systematically evaluating LLM robustness across diverse challenges.

The progression of robustness benchmarks reflects a growing understanding of LLM vulnerabilities and the need for comprehensive evaluation tools. AdvGLUE\cite{AdvGLUE}, was among the first to systematically evaluate adversarial robustness by applying 14 textual adversarial attack methods to GLUE tasks, revealing significant performance drops in state-of-the-art models and highlighting the necessity for more resilient architectures. BOSS\cite{BOSS} addressed out-of-distribution robustness by evaluating how models trained on specific distributions perform when encountering data from different distributions, emphasizing the importance of generalization in LLMs. IFEval\cite{IFEval} focused on instruction-following capabilities, providing a suite of tasks requiring models to adhere to specific instructions and measuring their ability to follow complex directives accurately. CIF-Bench\cite{CIF-Bench} extended this evaluation to multilingual contexts by assessing instruction-following in Chinese, testing models' zero-shot generalizability and highlighting challenges in cross-lingual understanding. PromptRobust\cite{PromptRobust} investigated the impact of prompt variations on model outputs, demonstrating that minor changes in prompt wording can significantly affect performance, thus underscoring the need for prompt-invariant models. RoTBench\cite{RoTBench} explored robustness in tool-use scenarios by assessing how models perform in environments with varying levels of noise and complexity, thereby evaluating their adaptability in real-world applications. Collectively, these benchmarks have expanded the evaluation landscape, moving beyond traditional accuracy metrics to encompass robustness against a spectrum of perturbations and challenges. Future directions involve developing standardized evaluation protocols, creating benchmarks for additional languages and modalities, and integrating robustness assessments into the model development lifecycle to ensure the deployment of reliable and trustworthy LLMs.

\subsubsection{Data Leak}

The widespread deployment of Large Language Models (LLMs) has raised significant concerns regarding data leakage, particularly the inadvertent disclosure of sensitive information such as Personally Identifiable Information (PII). This issue stems from the extensive pretraining of LLMs on vast corpora, which may include sensitive data, leading to the models memorizing and potentially reproducing such information during inference. The problem is exacerbated when models are fine-tuned on domain-specific datasets containing confidential information, increasing the risk of privacy breaches. Consequently, evaluating and mitigating data leakage has become a critical area of research.

Recent efforts have introduced multiple benchmarks aimed at systematically measuring data leakage risks in LLMs. For example, WikiMIA\cite{WikiMIA} focuses on monolingual data leakage by assessing classification performance using the minimum-k\% probability as a leakage indicator, operating over open datasets and including PII. KoLA\cite{KoLA} expands this analysis to bilingual contexts and evaluates generative models based on accuracy, relying on a hybrid dataset while excluding explicit PII. C$^2$LEVA\cite{C2LEVA} provides a large-scale benchmark also in a bilingual setting, integrating both classification and generation tasks. It uses mean win rate as the primary metric to measure leakage and includes PII within its evaluation scope.

These benchmarks reflect a growing recognition of the multifaceted nature of privacy risks in LLMs and highlight the necessity of evaluating models beyond traditional performance metrics. Data leakage in these contexts can be categorized by dimensions such as leakage rate, the tendency of a model to expose PII, and the ability of a model to detect and manage sensitive data. These dimensions inform the development of specialized evaluation protocols tailored to different model behaviors and data sources.

Collectively, these benchmarks have expanded the evaluation landscape, moving beyond traditional accuracy-oriented assessments to encompass privacy concerns associated with LLMs. Future directions involve developing standardized evaluation protocols, extending benchmarks to support more languages and modalities, and integrating privacy assessments throughout the model development lifecycle to ensure the deployment of reliable and trustworthy LLMs.

\subsubsection{Summary and Future Directions}

As large language models (LLMs) transition from research prototypes to real-world deployment in high-stakes domains, their safety and reliability have emerged as core concerns. Modern LLM evaluation frameworks have moved beyond capability-centric metrics to focus on systematically identifying and quantifying risks such as hallucinations, bias, adversarial vulnerabilities, and data leakage. Through fine-grained benchmarks spanning safety, hallucination, robustness, and privacy, researchers have developed multi-task, multilingual, and scenario-rich tools to empirically assess whether a model is reliable, safe, and trustworthy enough for real-world applications. These efforts collectively form the empirical foundation for liability-aware and secure LLM deployment.

Future directions in LLM risk assessment will center on several key areas. First, the development of unified evaluation frameworks is essential to integrate diverse risk dimensions and ensure composability and comparability across benchmarks. Second, the field must expand beyond English-centric evaluations to support low-resource languages and diverse cultural contexts. Third, new challenges such as long-context coherence, multi-turn consistency, and up-to-date knowledge integration call for more sophisticated evaluation protocols. Fourth, the growing complexity of attack strategies highlights the need for interactive adversarial modeling and continuous red-teaming. Fifth, \textbf{dynamic real-time evaluation} is becoming increasingly important—LLMs must demonstrate their ability to respond accurately to time-sensitive and evolving information, particularly in domains like finance, healthcare, and current events. Finally, privacy auditing and redaction capabilities must be embedded across the training and inference pipeline, pushing LLM safety evaluations from the model level to the system level.

\subsection{Agent} 
\begin{table}[h!]
\centering
\renewcommand{\arraystretch}{1}
\resizebox{\textwidth}{!}{
\begin{tabular}{llllllllll}
\hline
Benchmark & Focus & Language & Source & Data Type & Eval. & Indicators & Amount & Method & Citations \\ 
\hline
\multicolumn{10}{l}{\textbf{\textit{Specific Capability Assessment}}} \\
FlowBench \cite{FlowBench} & \begin{tabular}[t]{@{}l@{}} Workflow-Guided \\ Planning \end{tabular} & Monolingual & Hybrid & Generation & Hybrid & \begin{tabular}[t]{@{}l@{}}P; R; F1-score; \\ Success Rate, etc. \end{tabular} & 5313 & No & 16 \\
Robotouille \cite{Robotouille} & \begin{tabular}[t]{@{}l@{}} Asynchronous \\ Planning \end{tabular} & Monolingual & Human & Generation & AE & Success Rate & 300 & No & 5 \\
LLF-Bench \cite{LLF-Bench} & \begin{tabular}[t]{@{}l@{}} Learning from \\ Language Feedback \end{tabular} & Monolingual & Hybrid & Hybrid & AE & \begin{tabular}[t]{@{}l@{}} Reward; \\ Success Rate \end{tabular} & 8 sets & No & 17 \\
Mobile-Bench \cite{Mobile-Bench} & \begin{tabular}[t]{@{}l@{}} SmartPhone \\ Control \end{tabular} & Monolingual & Hybrid & Generation & AE & \begin{tabular}[t]{@{}l@{}}CheckPoint; \\ PassRate, etc. \end{tabular} & 832 & No & 36 \\
Spa-Bench \cite{Spa-Bench} & \begin{tabular}[t]{@{}l@{}} SmartPhone \\ Control \end{tabular} & Bilingual & Human & Generation & Hybrid & \begin{tabular}[t]{@{}l@{}}Success Rate; \\ Step Ratio, etc. \end{tabular} & 340 & Yes & 20 \\
BrowseComp \cite{BrowseComp} & Web Browse & Monolingual & Hybrid & Generation & AE & \begin{tabular}[t]{@{}l@{}} Accuracy; \\ Calibration Error \end{tabular} & 1226 & No & 19 \\
WebWalkerQA \cite{WebWalkerQA} & Web Browse & Bilingual & Hybrid & Generation & AE & \begin{tabular}[t]{@{}l@{}} Accuracy; \\ Action Count \end{tabular} & 680 & Yes & 6 \\
MultiAgentBench \cite{MultiAgentBench} & \begin{tabular}[t]{@{}l@{}} Collaboration \& \\ Competition \end{tabular} & Monolingual & Hybrid & Generation & LLM & \begin{tabular}[t]{@{}l@{}} KPI; \\ Planning Score, etc. \end{tabular} & 600 & Yes & 13 \\
MAgIC \cite{MAgIC} & Competition & Monolingual & Hybrid & Generation & AE & \begin{tabular}[t]{@{}l@{}} Win Rate; \\ Judgement, etc. \end{tabular} & 103 & Yes & 53 \\
ZSC-Eval \cite{ZSC-Eval} & \begin{tabular}[t]{@{}l@{}} Zero-shot \\ Coordination \end{tabular} & Monolingual & Model & Classification & Hybrid & \begin{tabular}[t]{@{}l@{}} BR-Prox;\\ BR-Div, etc. \end{tabular} & 2 envs. & Yes & 14 \\
\hline
\multicolumn{10}{l}{\textbf{\textit{Integrated Capability Assessment}}} \\
SmartPlay \cite{SmartPlay} & Game & Monolingual & Open Data & Classification & AE & \begin{tabular}[t]{@{}l@{}} Completion Rate;\\ Reward, et al. \end{tabular} & 6 games & No & 80 \\
BALROG \cite{BALROG} & Game & Monolingual & Open Data & Classification & AE & Average Progress & 6 games & No & 27 \\
\begin{tabular}[t]{@{}l@{}}Embodied Agent \\ Interface \cite{EmbodiedAgentInterface} \end{tabular} & \begin{tabular}[t]{@{}l@{}}Embodied Decision \\ Making \end{tabular} & Monolingual & Open Data & Generation & AE & \begin{tabular}[t]{@{}l@{}} F1; \\ Success Rate, etc. \end{tabular} & 438 & Yes & 67 \\
$\tau$-bench \cite{tau-bench} & \begin{tabular}[t]{@{}l@{}} Tool-Agent-User \\ Interaction \end{tabular} & Monolingual & Hybrid & Generation & AE & Pass@k & 165 tasks & No & 58 \\
TravelPlanner \cite{TravelPlanner} & Travel Planning & Monolingual & Hybrid & Generation & AE & \begin{tabular}[t]{@{}l@{}} Delivery Rate; \\ 3 Pass Rates \end{tabular} & 1225 & No & 170 \\
GAIA \cite{GAIA} & \begin{tabular}[t]{@{}l@{}} General AI \\ Assistants \end{tabular} & Monolingual & Human & Generation & AE & Exact Match & 466 & No & 185 \\
AgentQuest \cite{AgentQuest} & \begin{tabular}[t]{@{}l@{}} Agent \\ Improvement \end{tabular} & Monolingual & Open Data & Generation & AE & \begin{tabular}[t]{@{}l@{}} Progress Rate; \\ Repetition Rate \end{tabular} & 4 datasets & No & 17 \\
ColBench \cite{ColBench} & \begin{tabular}[t]{@{}l@{}}Backend Programming \\ \& Frontend Design \end{tabular} & Monolingual & Hybrid & Generation & AE & \begin{tabular}[t]{@{}l@{}} Tests Passed; \\ Success Rate, etc. \end{tabular} & 20K+ & Yes & 18 \\
AgentBench \cite{AgentBench} & Code, Game, Web & Monolingual & Hybrid & Generation & AE & \begin{tabular}[t]{@{}l@{}} Success Rate; \\ F1; Reward, etc. \end{tabular} & 1360 & No & 215 \\
AgentBoard \cite{AgentBoard} & \begin{tabular}[t]{@{}l@{}} Web, Tool, \\ Embodied AI, Game \end{tabular} & Monolingual & Hybrid & Generation & Hybrid & \begin{tabular}[t]{@{}l@{}} Progress Rate; \\ Success Rate, etc. \end{tabular} & 1013 & No & 17 \\
CharacterEval \cite{CharacterEval} & Role Playing & Monolingual & Open Data & Generation & LLM & \begin{tabular}[t]{@{}l@{}} Fluency; \\ Coherency; \\ Consistency, etc. \end{tabular} & 11376 & No & 71 \\
\hline
\multicolumn{10}{l}{\textbf{\textit{Domain Proficiency Evaluation}}} \\
TheAgentCompany \cite{TheAgentCompany} & \begin{tabular}[t]{@{}l@{}} Digital Software \\ Worker \end{tabular} & Monolingual & Hybrid & Generation & Hybrid & \begin{tabular}[t]{@{}l@{}} Completion Score; \\ Number of steps, etc. \end{tabular} & 175 & No & 39 \\
OSWorld \cite{OSWorld} & Computer Operation & Monolingual & Human & Generation & AE & Success Rate & 369 & No & 184 \\
Tapilot-Crossing \cite{Tapilot-Crossing} & \begin{tabular}[t]{@{}l@{}} Interactive \\ Data Analysis \end{tabular} & Monolingual & Hybrid & Hybrid & AE & \begin{tabular}[t]{@{}l@{}} Accuracy; \\ AccR \end{tabular} & 1024 & Yes & 11 \\
ScienceAgentBench \cite{ScienceAgentBench} & \begin{tabular}[t]{@{}l@{}} Data-driven \\ Scientific Discovery \end{tabular} & Monolingual & Hybrid & Generation & Hybrid & \begin{tabular}[t]{@{}l@{}} Valid Execution Rate; \\ Success Rate, etc. \end{tabular} & 102 & No & 53 \\
SciReplicate-Bench \cite{SciReplicate-Bench} & \begin{tabular}[t]{@{}l@{}} Algorithmic \\ Reproduction \end{tabular} & Monolingual & Open Data & Generation & AE & \begin{tabular}[t]{@{}l@{}} Execution Accuracy; \\ CodeBLEU, etc. \end{tabular} & 100 & Yes & 7 \\
MLGym-Bench \cite{MLGym-Bench} & AI Research & Monolingual & Open Data & Generation & AE & \begin{tabular}[t]{@{}l@{}} AUP Score; \\ Accuracy, etc. \end{tabular} & 13 tasks & Yes & 21 \\
InvestorBench \cite{InvestorBench} & \begin{tabular}[t]{@{}l@{}} Financial \\ Decision Making \end{tabular} & Monolingual & Open Data & Hybrid & AE & \begin{tabular}[t]{@{}l@{}} Cumulative Return; \\ Sharpe Ratio, etc. \end{tabular} & 3 tasks & No & 13 \\
BixBench \cite{BixBench} & \begin{tabular}[t]{@{}l@{}} Biological \\ Data Analysis \end{tabular} & Monolingual & Hybrid & Hybrid & Hybrid & Accuracy & 296 & No & 6 \\
AgentClinic \cite{AgentClinic} & \begin{tabular}[t]{@{}l@{}} Clinical \\ Decision Making \end{tabular} & Monolingual & Open Data & Generation & LLM & \begin{tabular}[t]{@{}l@{}} Diagnostic Accuracy; \\ Confidence, etc. \end{tabular} & 1544 & No & 69 \\
CourtBench \cite{CourtBench} & Legal Reasoning & Monolingual & Hybrid & MCQA & AE & Accuracy & 124 & Yes & 11 \\
\hline
\multicolumn{10}{l}{\textbf{\textit{Safety \& Risk Evaluation}}} \\
ASB \cite{ASB} & Attacks \& Defenses & Monolingual & Hybrid & Generation & AE & \begin{tabular}[t]{@{}l@{}} Attack Success Rate; \\  Refuse Rate, etc. \end{tabular} & 50 tasks & Yes & 49 \\
AgentHarm \cite{AgentHarm} & Jailbreak Attacks & Monolingual & Hybrid & Generation & Hybrid & \begin{tabular}[t]{@{}l@{}} Harm Score; \\ Refusal Rate, etc. \end{tabular} & 440 & No & 67 \\
SafeAgentBench \cite{SafeAgentBench} & \begin{tabular}[t]{@{}l@{}} Safe Task \\ Planning \end{tabular} & Monolingual & Model & Generation & Hybrid & \begin{tabular}[t]{@{}l@{}} Rejection Rate; \\ Success Rate, etc. \end{tabular} & 750 & No & 6 \\
R-Judge \cite{R-Judge} & \begin{tabular}[t]{@{}l@{}} Safety Risk \\ Awareness \end{tabular} & Monolingual & Hybrid & Hybrid & Hybrid & \begin{tabular}[t]{@{}l@{}} Safety Judgment; \\ Risk Identification \end{tabular} & 569 & No & 101 \\
\hline
\end{tabular}
}
\caption{Summary of representative benchmarks for LLM agent. Evaluation methods are abbreviated as AE (Automated Evaluation), ME (Mixed Evaluation). The 'Method' column indicates if the paper proposed a new methodology (Yes/No).}
\label{tab:agent_summary}
\end{table}

LLM agents are autonomous systems built upon foundation large language models, designed to transcend static prompt-response interactions and engage in goal-driven behaviors. By integrating components such as planning modules, tool-use capabilities, memory systems, and observation loops, these agents can decompose complex objectives into actionable steps, interact dynamically with external environments, and iteratively adapt their strategies until task completion. As LLM agents find increasing application in real-world scenarios, establishing systematic and comprehensive evaluation methodologies becomes essential. As summarized in Table~\ref{tab:agent_summary}, this survey organizes the evaluation framework of LLM agents into four key dimensions: (1) specific capability assessment, focusing on the fine-grained evaluation of individual functions (e.g., planning, reasoning, competition) and execution abilities (e.g., tool use, external control); (2) integrated capability assessment, emphasizing the coordination and synergy of multiple abilities in solving complex tasks; (3) domain proficiency evaluation, focusing on assessing the application of specialized knowledge and the effectiveness in performing tasks within specific professional domains; and (4) safety \& risk evaluation, focusing on agents’ resilience, susceptibility, and protective mechanisms in adversarial or unsafe scenarios.

\subsubsection{Specific Capability Assessment}
Assessing the specific capabilities of LLM-based agents is essential for understanding their functional reliability and task generalization limits. Instead of evaluating end-to-end task success alone, this line of research focuses on isolating and benchmarking core abilities such as planning, reasoning, tool use, and interactive behavior. Such evaluations enable more interpretable diagnostics and drive progress on targeted weaknesses in agent design.

A series of benchmarks center on planning and reasoning abilities. FlowBench~\cite{FlowBench} evaluates how LLM agents leverage workflow knowledge to perform structured, domain-specific planning. Robotouille~\cite{Robotouille} targets asynchronous planning, requiring agents to handle delayed effects and overlapping actions in long-horizon tasks. LLF-Bench~\cite{LLF-Bench} focuses on an agent’s ability to improve through iterative language feedback, testing whether agents can learn across turns using naturalistic supervision. WebWalkerQA~\cite{WebWalkerQA} examines reasoning over hierarchical web structures, evaluating how agents navigate multi-layered pages to extract complex information.

Another major direction is evaluating external control and tool-use capabilities. SPA-Bench~\cite{Spa-Bench} and Mobile-Bench~\cite{Mobile-Bench} assess LLM agents’ performance on mobile device control, with tasks spanning single-app actions to multi-app collaboration. These benchmarks emphasize interface understanding, execution reliability, and reasoning under UI constraints. BrowseComp~\cite{BrowseComp} extend this to web environments, measuring agents’ effectiveness in retrieving hard-to-find facts through persistent, tool-mediated browsing.

Multi-agent scenarios test coordination, competition, and social reasoning. MultiAgentBench~\cite{MultiAgentBench} examines agent collaboration across different topologies and tasks, while MAgIC~\cite{MAgIC} incorporates social deduction games and game-theoretic settings to probe deception, self-awareness, and judgment. ZSC-Eval~\cite{ZSC-Eval} introduces the challenge of zero-shot coordination, where agents must generalize to novel partners in cooperative environments, with evaluation grounded in behavior diversity and robustness.

These benchmarks emphasize that evaluating isolated cognitive or interaction abilities is foundational for building trustworthy LLM agents. By pinpointing weaknesses in planning, reasoning, tool use, and multi-agent dynamics, specific capability assessment provides fine-grained insights that inform more robust and modular agent design.

\subsubsection{Integrated Capability Assessment}
Integrated capability assessment focuses on evaluating how well LLM agents coordinate multiple abilities—such as reasoning, planning, tool use, memory, and interaction—to solve complex, multi-step tasks. Unlike evaluations targeting isolated skills, this assessment emphasizes holistic competence in dynamic environments, where agents must sequence decisions, adapt to feedback, follow constraints, and operate across multiple modalities. It reflects the practical readiness of LLM agents to handle real-world challenges that demand cognitive integration and situational flexibility.

Several benchmarks assess agents’ integrated reasoning and decision-making in game-like or embodied environments. SmartPlay~\cite{SmartPlay} uses diverse games like Tower of Hanoi and Minecraft to decompose and assess nine essential agent abilities, including object dependency reasoning and learning from history. BALROG~\cite{BALROG} evaluates both LLMs and VLMs in complex games requiring planning, spatial reasoning, and exploration across various difficulty levels. Embodied Agent Interface~\cite{EmbodiedAgentInterface} provides a unified framework for embodied tasks and evaluates agents across subtasks like goal interpretation, action sequencing, and transition modeling.

Some benchmarks simulate real-world task environments requiring tool usage, constraint tracking, and user interaction. $\tau$-bench~\cite{tau-bench} evaluates agents in multi-turn tool-augmented dialogues with domain-specific rules, measuring goal satisfaction and behavior consistency. TravelPlanner~\cite{TravelPlanner} tests complex real-world planning through a large-scale travel sandbox, assessing tool use, information integration, and constraint handling. GAIA~\cite{GAIA} offers questions requiring general web search, multimodal perception, and robust reasoning, targeting human-level generalist performance. AgentQuest~\cite{AgentQuest} introduces a modular benchmarking framework with extensible metrics to diagnose agent progress and failure modes over multi-step tasks.

Multi-turn interaction and collaborative reasoning are also critical to integrated capability assessment. SWEET-RL~\cite{ColBench} focuses on multi-turn collaboration tasks like frontend design, using a step-wise reward system to optimize agent behavior. ColBench~\cite{ColBench}, built alongside it, facilitates realistic back-and-forth problem solving with human collaborators. AgentBench~\cite{AgentBench} and AgentBoard~\cite{AgentBoard} provide large-scale evaluation environments across multiple domains, incorporating multi-turn planning, decision-making, and error tracking to reveal bottlenecks in agent performance. CharacterEval~\cite{CharacterEval} adds a role-playing dimension to agent evaluation in Chinese, testing coherence, personality consistency, and long-term conversation management.

Together, these benchmarks emphasize the necessity of evaluating LLM agents’ coordination and synergy across multiple core abilities to solve complex, multi-step problems. By focusing on holistic competence in dynamic and interactive environments, integrated capability assessments provide deeper insights into agents’ real-world readiness, exposing challenges in adaptability, multi-modal reasoning, and sustained interaction.

\subsubsection{Domain Proficiency Evaluation}
As language agents transition from general-purpose tools to specialized assistants, evaluating their domain proficiency becomes essential. Unlike general reasoning tasks, these evaluations focus on the agent’s ability to apply expert knowledge, follow domain-specific procedures, and complete professional tasks with precision. Such tasks often involve high-stakes decision-making, intricate workflows, multimodal inputs, or specialized tools, requiring agents to go beyond generic language understanding.

To this end, a wide range of domain-specific benchmarks have been proposed. In workplace and productivity scenarios, TheAgentCompany~\cite{TheAgentCompany} assesses agents on realistic office tasks such as browsing, coding, and intra-team communication, while OSWorld~\cite{OSWorld} provides a scalable, interactive operating environment that evaluates agents’ ability to perform open-ended computer tasks across platforms. In data science and scientific research, Tapilot-Crossing~\cite{Tapilot-Crossing} benchmarks interactive data analysis capabilities, ScienceAgentBench~\cite{ScienceAgentBench} and SciReplicate-Bench~\cite{SciReplicate-Bench} evaluate agents’ ability to generate and reproduce scientific code, and MLGym-Bench~\cite{MLGym-Bench} focuses on end-to-end AI research tasks from hypothesis generation to model evaluation.

The financial domain is covered by InvestorBench~\cite{InvestorBench}, which evaluates agents across diverse financial instruments and market scenarios, while the biomedical and healthcare domains are represented by BixBench~\cite{BixBench} and AgentClinic~\cite{AgentClinic}, which test agents on long-horizon bioinformatics tasks and clinical decision-making under multimodal constraints, respectively. 
In legal contexts, CourtBench~\cite{CourtBench} serves as the evaluation suite within AgentCourt, measuring legal reasoning, cognitive agility, and argumentative rigor of adversarially evolved lawyer agents in simulated courtroom trials. 
For software engineering and algorithm comprehension, SciReplicate-Bench~\cite{SciReplicate-Bench} further challenges agents to extract, understand, and implement methods from recent NLP papers.

Collectively, these benchmarks underscore the gap between general language proficiency and domain-specific expertise. They highlight that while current agents demonstrate partial competence, substantial limitations remain in reliability, depth of understanding, and tool integration across professional settings. As such, domain proficiency evaluation plays a critical role in driving the development of trustworthy, capable, and specialized LLM agents.

\subsubsection{Safety \& Risk Evaluation}
Safety and risk evaluation addresses the robustness of LLM agents under adversarial, malicious, or failure-prone conditions. As agents move beyond static text generation to tool use, memory manipulation, and autonomous decision-making, the surface for potential misuse and failure grows significantly. This evaluation dimension focuses on whether agents can maintain aligned, reliable behavior while resisting manipulation, detecting hazards, and handling unsafe instructions. It plays a crucial role in ensuring trustworthy deployment, especially in sensitive or high-stakes domains.

One category of benchmarks targets adversarial vulnerabilities and attack-resistance. Agent Security Bench (ASB)~\cite{ASB} provides a comprehensive framework covering a wide range of real-world scenarios, agents, and attack types. It reveals significant weaknesses across agent components, including prompt processing, memory, and tool interfaces. AgentHarm~\cite{AgentHarm} focuses on agent misuse by introducing harmful task prompts across domains such as fraud and cybercrime. It finds that many leading agents comply with malicious requests, and simple jailbreak templates can induce harmful multi-step behaviors.

A second line of work explores safety in embodied and task-execution environments. SafeAgentBench~\cite{SafeAgentBench} evaluates whether agents can recognize and avoid hazardous instructions in interactive simulations. Results show that current agents generally lack the ability to reject unsafe plans, even when task understanding is adequate. R-Judge~\cite{R-Judge} shifts the focus to risk judgment, assessing whether agents can identify safety issues from interaction records. While some frontier models show moderate awareness, overall performance remains far from robust, highlighting the difficulty of safety reasoning in open-ended settings.

Collectively, these benchmarks reveal that LLM agents remain highly vulnerable to both adversarial manipulation and operational hazards. Improving their safety requires deeper integration of risk modeling, hazard detection, and behavioral safeguards, moving beyond surface-level refusals toward genuine situational awareness and robustness.

\subsubsection{Summary and Future Directions}
The evaluation of LLM agents has evolved into a multi-dimensional endeavor encompassing specific capabilities, integrated competencies, domain proficiency, and safety robustness. 
Recent benchmarks have illuminated the growing sophistication of agents across planning, reasoning, tool use, and interaction, while also highlighting key limitations in cross-skill coordination, domain-specific expertise, and adversarial resilience. 
Specific capability assessments provide fine-grained diagnostics of isolated functions; integrated evaluations reflect holistic problem-solving in dynamic environments; domain benchmarks underscore the gap between general language ability and professional-grade performance; and safety evaluations expose vulnerabilities in real-world scenarios. 
Overall, this emerging ecosystem of benchmarks offers a systematic lens through which to assess the functional maturity and deployment readiness of LLM agents.

Looking ahead, future research may prioritize three interlinked directions. 
First, there is a pressing need for evaluation compositionality, namely developing unified frameworks that assess how well agents orchestrate diverse skills across evolving contexts, rather than excelling in isolated tests. 
Second, grounded and continuous evaluation must be strengthened, where agent behavior is tested in realistic, long-term deployments involving dynamic feedback loops, imperfect tools, and human collaboration. 
Third, robustness and safety must move beyond binary refusal detection toward deeper modeling of intent, risk context, and adaptive safeguards. 
Achieving trustworthy autonomy will require benchmarks that are not only broader in coverage but also richer in interpretability and diagnostic precision, enabling the next generation of LLM agents to act reliably in complex, open-ended, and high-stakes environments.

\subsection{Others}
\begin{table}[h!]
\centering
\renewcommand{\arraystretch}{1}
\resizebox{\textwidth}{!}{
\begin{tabular}{llllllllll}
\hline
Benchmark & Focus & Language & Source & Data type & Eval. & Indicators & Amount & Method & Citations \\ \hline
PET-BENCH \cite{pet-bech} & Virtual Pet Companionship & Monolingual &Hybrid&Hybrid&AE& \begin{tabular}[t]{@{}l@{}}BLEU, ROUGE, Accuracy\end{tabular} &7,815&No&0\\
TP-RAG \cite{tp-rag}& Travel Planning & Monolingual &Hybrid&Generation&AE&\begin{tabular}[t]{@{}l@{}}FR, RR, DMR, STR\end{tabular} &2,348&No&1\\
FLUB \cite{FLUB}& Fallacy Understanding & Monolingual &Hybrid&Hybrid&AE&\begin{tabular}[t]{@{}l@{}}Accuracy, F-1 Score\\GPT-4 Score\end{tabular} &834&No&15\\
CDEval~\cite{cdeval}& Cultural Dimensions&Multilingual &Hybrid&MCQA&AE&Average Likelihood &2,953&No&22\\
NORMAD-ETI~\cite{normad}&Cultural Adaptability & Monolingual &Hybrid&Classification&AE&Accuracy&2.6k&No&6\\
EmotionQueen~\cite{emotionqueen}& Emotional Intelligence&Monolingual & Model&Generation&AE&\begin{tabular}[t]{@{}l@{}}Pass Rate\\Win Rate\end{tabular} &10000&No&69\\
OR-Bench~\cite{orbench}&Over-refusal&Monolingual&Hybrid&Generation&AE& \begin{tabular}[t]{@{}l@{}}Over-refusal Rate, Acceptance Rate\end{tabular}&80,000&No&79\\
SocialStigmaQA~\cite{SocialStigmaQA}&Stigma Amplification & Monolingual &Hybrid&MCQA&AE&Biased Answer Percent &10k&No&22\\
Shopping MMLU~\cite{ShoppingMMLU} &Online Shopping &Multilingual &Web&Hybrid&AE&\begin{tabular}[t]{@{}l@{}}Hit Rate@3, NDCG, Micro F1\\ROUGE-L, BLEU\end{tabular} &14683&No&17\\
JudgeBench~\cite{JudgeBench}& LLM-based Judges &Monolingual &Hybrid&MCQA&AE&\begin{tabular}[t]{@{}l@{}}Accuracy\end{tabular} &350&No&71\\
LLM-Evolve~\cite{LLM-Evolve}& Evolving Capability& Monolingual&Opendata Extend&MCQA, Generation&AE&\begin{tabular}[t]{@{}l@{}} Success Rate, F1, Reward ,Game Progress,\\Step Success Rate, Accuracy\end{tabular} &23k&No&9\\
SUC~\cite{tablellm}&Structural Understanding Capabilities& Monolingual&Opendata&Generation&AE&\begin{tabular}[t]{@{}l@{}}Accuracy\end{tabular} &255k&No&206\\
DOCBENCH~\cite{DOCBENCH}&Document reading & Monolingual&Hybrid&MCQA&AE&\begin{tabular}[t]{@{}l@{}}Accuracy\end{tabular} &1102&No&13\\
ROUTERBENCH~\cite{RouterBench}&Routers Hinders Progress  & Monolingual&Hybrid&Generation&AE&\begin{tabular}[t]{@{}l@{}}Exact Match, GPT-4, Accuracy,\\ Total Cost (Dollars)\end{tabular} &405,467&Yes&81\\
GAME COMPETITIONS~\cite{Grid-Based-Game}&Game Simulation& Monolingual&Hybrid&Generation&AE&\begin{tabular}[t]{@{}l@{}}Win Rate,\\ Draw Rates, Disqualification
Rate\end{tabular} &2310&No&16\\
RTLLM 2.0~\cite{OpenLLM-RTL}&Design RTL generation  & Monolingual&Human&Generation&AE&\begin{tabular}[t]{@{}l@{}}Pass@k\end{tabular} &50&No&16\\
AD-LLM~\cite{AD-LLM}& Anomaly Detection& Monolingual&Opendata&Classification&AE&\begin{tabular}[t]{@{}l@{}}AUROC, AUPRC\end{tabular} &190k&Yes&16\\
ZIQI-Eval~\cite{ZIQI-Eval}&Musical Abilities &Monolingual &Human&MCQA&AE&\begin{tabular}[t]{@{}l@{}}Precision, Recall, Accuracy, F1\end{tabular} &14,000&No&15\\
VisEval~\cite{VisEval}&Tabular Data Vsualization& Monolingual&Hybrid&Generation&AE&\begin{tabular}[t]{@{}l@{}}Validity, Legality, Readability Rating\end{tabular} & 2,524&No&44\\
\hline
\end{tabular}
}
\caption{Summary of other representative others benchmarks. The 'Method' column indicates if the paper proposed a new methodology (Yes/No).}
\label{tab:other_benchmarks}
\end{table}

In recent years, an increasing body of research has shifted toward evaluating large language models (LLMs) from unique, human-centered perspectives, moving beyond conventional NLP tasks such as question answering, summarization, and translation. These evaluations extend into more complex domains including cultural adaptability, emotional intelligence, value alignment, real-world task execution, and multimodal technical capabilities. 

In the realm of cultural intelligence and social adaptability, CDEval~\cite{cdeval} conducts multilingual multiple-choice testing across six cultural dimensions and seven domains with 2,953 samples, while NORMAD-ETI~\cite{normad} uses 2.6k classification tasks from 75 cultural contexts to assess narrative adaptability. SocialStigmaQA~\cite{SocialStigmaQA} examines bias amplification on 93 social stigma topics through 10k multiple-choice samples, revealing that models may still exhibit systematic tendencies toward certain groups even under explicit prompting.

Emotional understanding and interpersonal interaction form another core dimension. EmotionQueen~\cite{emotionqueen} evaluates the recognition of implicit emotions and appropriateness of affective responses in 10k generation tasks, using PASS and WIN rates as metrics. Similarly, PET-Bench~\cite{pet-bech} tests emotional support and memory-consistent dialogue through 7,815 virtual pet companionship scenarios, combining BLEU, ROUGE, and accuracy. Complementary to these are studies on value alignment, reasoning robustness, and safety balance. OR-Bench~\cite{orbench} analyzes 80k over-refusal cases to measure the balance between refusals and acceptances; FLUB~\cite{FLUB} tests logical fallacy recognition over 834 samples with accuracy, F1, and GPT-4 scores; JudgeBench~\cite{JudgeBench} assesses the consistency of LLM-based judges across 350 multiple-choice questions.

Simulated real-world tasks provide more application-oriented evaluation settings. Shopping MMLU~\cite{ShoppingMMLU} measures reasoning, behavior modeling, and recommendation performance in e-commerce scenarios using 14,683 multilingual hybrid tasks, with metrics including hit rate@3, NDCG, micro-F1, ROUGE-L, and BLEU. TP-RAG~\cite{tp-rag} evaluates 2,348 travel planning tasks on itinerary design, destination matching, and scheduling accuracy. DOCBENCH~\cite{DOCBENCH} contains 1,102 document reading comprehension tasks; LLM-Evolve~\cite{LLM-Evolve} tracks capability evolution over 23k extended tasks; SUC~\cite{tablellm} features 255k structured data understanding tasks assessed by accuracy; and VisEval~\cite{VisEval} evaluates 2,524 table visualization outputs for validity, legality, and readability.

Several benchmarks explore technical and multimodal extensions. ROUTERBENCH~\cite{RouterBench} evaluates routing strategies over 405,467 tasks with exact match, GPT-4 scores, accuracy, and economic cost. GAME COMPETITIONS~\cite{Grid-Based-Game} measure strategic reasoning through win, draw, and disqualification rates in 2,310 game simulations. RTLLM 2.0~\cite{OpenLLM-RTL} targets RTL code generation for hardware design with pass@k evaluation over 50 hand-crafted tasks. AD-LLM~\cite{AD-LLM} tests anomaly detection on 190k classification tasks using AUROC and AUPRC. ZIQI-Eval~\cite{ZIQI-Eval} assesses musical understanding and generation on 14k multiple-choice tasks, measuring precision, recall, accuracy, and F1.

Overall, these diverse benchmarks reflect a clear trend: LLM evaluation is transitioning from purely linguistic competence assessment toward multidimensional, human-centered, socially-aware, and application-driven evaluation. By encompassing cultural adaptability, emotional perception, value alignment, real-world task complexity, and technical extension, these works establish a methodological foundation for developing next-generation language models that are not only linguistically capable but also socially responsible and practically valuable.

\section{Conclusion}

This survey maps the evolving landscape of LLM evaluation benchmarks, drawing on a detailed analysis of 283 benchmarks organized into three core categories: general capabilities, domain-specific expertise, and target-specific functionalities. Tracing their progression from task-specific leaderboards to multidimensional frameworks, we highlight how these benchmarks have both reflected and driven advances in LLM capabilities, from foundational linguistics to domain mastery, and from standalone performance to safety-critical reliability.  
Our taxonomy reveals key tensions: general benchmarks often sacrifice depth for breadth; domain-specific assessments risk over-specialization; and target-specific metrics struggle to balance technical rigor with real-world relevance. These challenges intensify as LLMs operate in dynamic, multi-agent, high-stakes environments, where static datasets and single-turn metrics fail to capture emergent behaviors or societal impacts.  
As LLMs integrate into sociotechnical systems, evaluation must shift from measuring "what models can do" to "how they should perform responsibly." Future benchmarks require dynamism (to match model evolution), causality (to explain outcomes), inclusion (to avoid bias), and robustness (to anticipate risks). Achieving this requires cross-disciplinary collaboration to align technical rigor with societal values.

\newpage
{
    \small
    \bibliographystyle{unsrt}
    \bibliography{reference}

}
\end{document}